\documentclass[pdflatex,iicol,sn-mathphys-num]{sn-jnl}

\usepackage{graphicx}
\usepackage{multirow}
\usepackage{amsmath,amssymb,amsfonts}
\usepackage{amsthm}
\usepackage{mathrsfs}
\usepackage[title]{appendix}
\usepackage{xcolor}
\usepackage{textcomp}
\usepackage{manyfoot}
\usepackage{booktabs}
\usepackage{algorithm}
\usepackage{algorithmicx}
\usepackage{algpseudocode}
\usepackage{listings}

\usepackage{subcaption}
\usepackage{tabularx}
\usepackage{array}
\usepackage{float}
\usepackage{longtable}
\usepackage{bm}
\usepackage{xurl}
\usepackage{placeins}


\theoremstyle{thmstyleone}

\theoremstyle{thmstyletwo}

\newtheorem{remark}{Remark}

\theoremstyle{thmstylethree}

\begin{document}

\title[SplineNet]{SplineNet: An Isogeometric Deep Learning Method for Complex Shells}

\author[1]{\fnm{Shizhou} \sur{Luo}}\email{LuoShizhou@sjtu.edu.cn}

\author*[1]{\fnm{Xiaodong} \sur{Wei}}\email{xiaodong.wei@sjtu.edu.cn}

\affil[1]{\orgdiv{Global College}, 
\orgname{Shanghai Jiao Tong University}, 
\orgaddress{\city{Shanghai}, \postcode{200240}, \country{China}}}

\abstract{
We present a novel isogeometric deep learning method, termed \textit{SplineNet}, for the seamless design and analysis of shell structures with complex geometries. The proposed approach is built upon watertight spline representations, e.g., analysis-suitable unstructured T-splines, and features exact geometric descriptions of Computer-Aided Design (CAD) models in neural networks. B\'ezier extraction is used to build the network architecture, where Bernstein polynomials serve as the nonlinear activation functions. SplineNet can be applied in a data-free or data-driven way. In the data-free case, energy-based formulations can be naturally incorporated as loss terms, which fulfill the need of Computer-Aided Engineering (CAE) and can be accurately calculated. In particular, the Kirchhoff--Love (KL) model is adopted to solve for the mechanical behaviors of shell structures. This way, CAD and CAE can be tightly integrated in a deep neural network without the time-consuming model/data exchange process. In the data-driven case, SplineNet can be used as the trunk net of Deep Operator Networks (DeepONet) to provide interpretability. Given such a trained network and unseen input data, results can be immediately obtained without retraining the network or repeatedly performing the traditional workflow for analysis. In the end, a variety of numerical examples are studied to demonstrate the effectiveness of the proposed method, especially when real-world complex geometries are involved.
}

\keywords{Deep Operator Networks, Interpretability, Isogeometric Analysis, Analysis-suitable Unstructured Splines, B\'ezier Extraction}

\maketitle

\section{Introduction}
Isogeometric Analysis (IGA) was proposed to bridge the gap between Computer-Aided Design (CAD) and Computer-Aided Engineering (CAE) by employing spline-based representations used in CAD as basis functions for numerical analysis~\cite{cottrell2009isogeometric}. By directly utilizing exact geometric descriptions, IGA enables higher geometric fidelity and improved numerical performance compared with the finite element method (FEM)~\cite{reddy1993introduction}. In particular, spline-based discretizations possess higher-order continuity, making them especially suitable for high-order partial differential equations (PDEs), such as the Kirchhoff--Love (KL) shell formulation~\cite{kiendl2009isogeometric}. Despite these advantages, traditional IGA methods become computationally expensive for large-scale problems and require repeated simulations in design optimization~\cite{nagy2010isogeometric} and uncertainty quantification~\cite{beck2019iga}.

To overcome these drawbacks, leveraging deep learning to enhance the performance of traditional PDE solvers has attracted significant attention~\cite{wang2024artificial}. Based on learning objectives, these approaches can be broadly classified into two categories: solution learning and operator learning.

In the solution learning paradigm, the neural network acts as a function approximator to solve a specific problem. Many methods have been studied accordingly based on both strong and energy/weak forms of PDEs, such as physics-informed neural networks (PINNs)~\cite{raissi2019physics}, the deep Ritz method (DRM)~\cite{yu2018deep}, the deep energy method (DEM)~\cite{samaniego2020energy}, and variational physics-informed neural networks (VPINNs)~\cite{kharazmi2021hp}. PINNs utilize strong form residuals to define loss functions. They provide a flexible, mesh-free approach for solving forward and inverse problems. However, they require high-order derivatives and lead to training difficulties~\cite{basir2022investigating, chickering2024quasilinear}. DRM, DEM, and VPINNs employ the energy/weak form of PDEs as loss functions, which can help reduce the order of derivatives~\cite{fuhg2022mixed} and improve computational efficiency~\cite{bai2025energy}. Despite these improvements, in many implementations of these methods, essential boundary conditions are imposed through penalty terms that cause conflicting gradients during training~\cite{wang2021understanding, berrone2023enforcing}.

Another line of solution-learning methods is inspired by classical discretization techniques, such as hierarchical deep learning neural networks (HiDeNN)~\cite{saha2021hierarchical, zhang2021hierarchical, liu2023hidenn, lu2023convolution} and finite element neural network interpolation (FENNI)~\cite{vskardova2025finite, daby2025finite}. These methods construct finite-element-like shape functions using neural network architectures. Such networks can enhance interpretability, accurately impose boundary conditions, and facilitate advanced strategies such as \textit{r}-adaptivity~\cite{zhang2021hierarchical, vskardova2025finite} and multigrid training~\cite{vskardova2025finite}. However, these approaches remain instance-specific: any change in geometries, boundary conditions, or loadings requires retraining.

In contrast, operator learning methods aim to learn mappings from input functions to output solution fields, allowing them to provide instant predictions for unseen inputs without retraining. For example, Fourier neural operators (FNO)~\cite{li2020fourier} leverage the fast Fourier transform (FFT) to achieve efficient and resolution-independent operator learning. However, FNO typically requires functions to be represented on uniform grids, which limits its application to complex domains. Geo-FNO~\cite{li2023fourier} addresses this issue by learning to transform an irregular physical domain into a latent uniform grid. Graph kernel networks (GKNs)~\cite{li2020neural} approximate kernel integral operators on graph-based discretizations and can therefore handle irregular meshes. Deep Operator Networks (DeepONet)~\cite{lu2021learning} employ a branch net to encode input functions and a trunk net to encode output locations, providing a flexible framework for learning nonlinear operators. Several variants exist to enhance the performance of DeepONet. For example, the residual U-Net is leveraged to encode complex geometries~\cite{he2023novel}. Point-DeepONet~\cite{park2026point} integrates PointNet to learn geometric information from point clouds. Neural operators on Riemannian manifolds (NORM)~\cite{chen2024learning} represent the input and output using the eigenfunctions of the Laplace--Beltrami operator defined on the underlying manifold, thereby providing a spectral basis for learning the input-output mapping. These methods extend operator learning beyond regular grids.

However, there exists a fundamental gap for these methods to be applied in engineering design, because their geometric representations are recognized by sampled points in the computational domain rather than by CAD models. This inconsistency poses several limitations. First, the performance of neural nets may be sensitive to the choice of point locations, especially in problems where local features, such as stress concentration and cracks, are important~\cite{wu2023comprehensive, gu2023enriched, visser2026pacmann}. Second, even with a carefully curated point set, analysis results still need to be mapped back to the original CAD model to facilitate design iterations, which can be error-prone, especially for large-scale problems. Third, visualization of analysis results requires additional care, including the choice of point locations for evaluating the design and the triangulation of such points to yield intuitive and high-quality visualizations, which often require domain knowledge. Therefore, incorporating the isogeometric concept, i.e., using the same geometric representation everywhere, into learning-based methods has significant potential to enhance conventional numerical solvers while seamlessly integrating CAD and CAE in neural nets.

Recent studies have started to exploit this idea by integrating spline representations into solution learning methods. For instance, isogeometric neural networks (IGN)~\cite{gasick2023isogeometric} train a neural network to predict control variables, analogous to nodal values in FEM, rather than pointwise solution values, thereby reducing the computational burden and keeping the learned solution in an isogeometric approximation space. Isogeometric convolution HiDeNN (C-IGA)~\cite{zhang2023isogeometric} introduces convolution into HiDeNN to reproduce splines and support adaptive refinement. Later, it has been extended to the multi-patch case~\cite{zhang2025multi} by imposing compatibility conditions across patch interfaces. Alternatively, multi-patch isogeometric neural solvers~\cite{von2025multi} formulate neural ansatz functions on IGA reference domains, where interface conditions are enforced and an energy-based loss is used for training. Nonetheless, multi-patch representations are rather restrictive in modeling complex geometries, as they require a multi-block mesh structure that is generally difficult to achieve. Moreover, enforcing interface conditions can be tedious and challenging for high-order problems such as Kirchhoff--Love shells.

\begin{figure*}[tbp]
\centering
\includegraphics[width=0.9\linewidth]{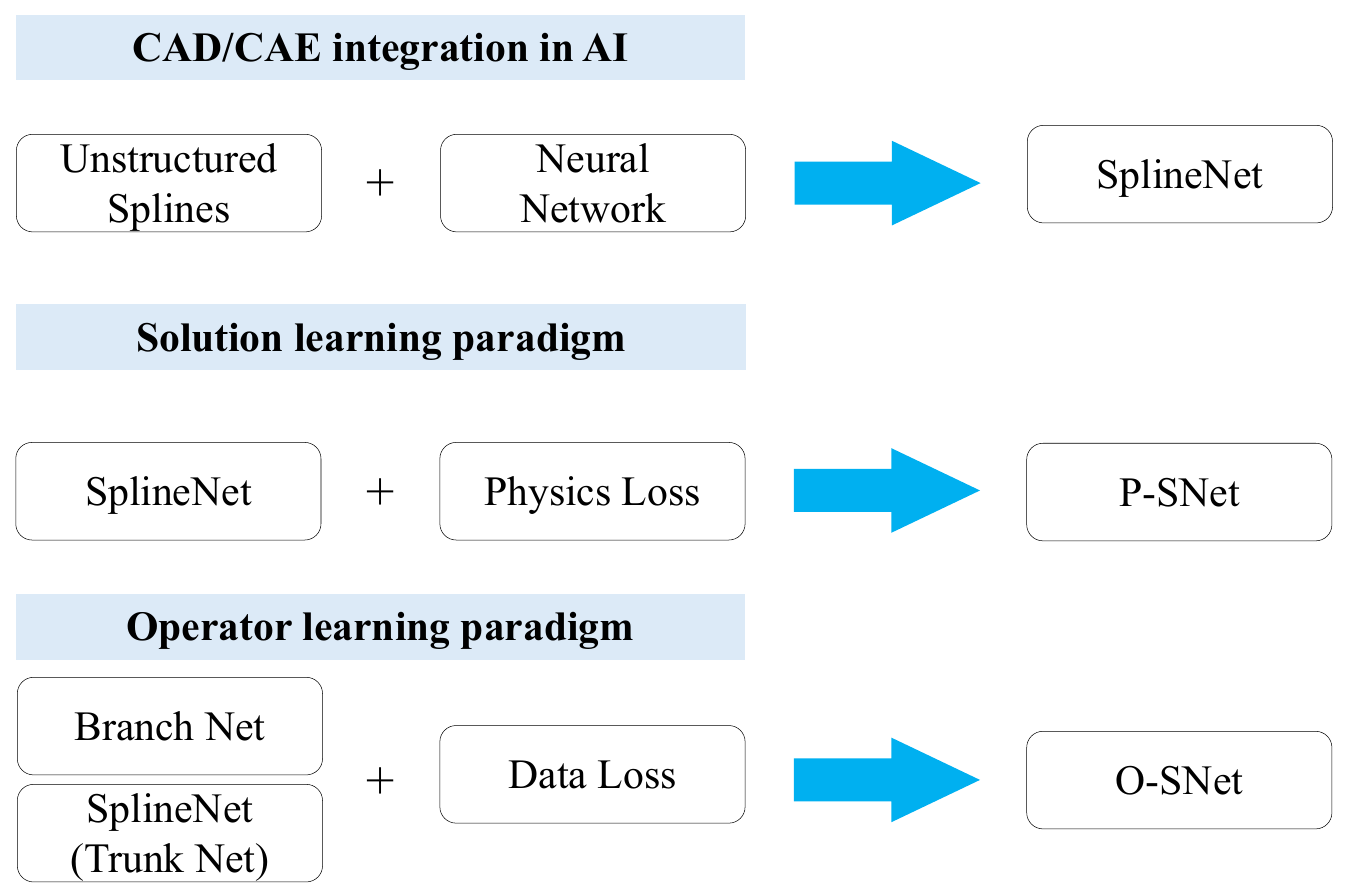}
\caption{Overview of SplineNet and its applications in both solution learning and operator learning. Unstructured splines are explicitly embedded into the neural network to construct the foundational SplineNet (top). It supports two distinct paradigms: a solution learning paradigm (P-SNet) driven by a physical loss (e.g., the total energy of Kirchhoff-Love shells) for solving PDEs, and an operator learning paradigm (O-SNet) that incorporates an additional branch net and a data loss to achieve instant predictions under varying input conditions.}
\label{overview}
\end{figure*}

To address these issues, we propose a novel isogeometric deep learning method, termed \textit{SplineNet}, for the seamless design and analysis of thin shell structures with complex geometries; see Fig.~\ref{overview} for an overview. 

Inspired by basis-function-aware networks, e.g., HiDeNN and C-IGA, SplineNet directly embeds analysis-suitable unstructured splines (ASUTS)~\cite{wei2022analysis}, an advanced spline technique for modeling complex geometries in a watertight manner, into neural networks via B\'ezier extraction. The weights and biases of SplineNet are obtained from the extraction matrices, with Bernstein polynomials serving as the activation functions. The inputs are parametric coordinates, whereas the learnable parameters are control variables or control points, depending on the application of interest. In this way, the geometric model of SplineNet remains the same in both design and analysis. Due to the modeling capability of ASUTS, SplineNet can accommodate geometries from real-world applications that go beyond simple single- or multi-patch settings.

SplineNet supports both solution learning and operator learning. In the former case, SplineNet acts as a physics-informed neural net to solve governing PDEs using an energy-based method for a single instance, and it is referred to as \textit{Physics-informed SplineNet} (\textit{P-SNet}). In the latter case, the trunk net of DeepONet is replaced with SplineNet. As a result, the solution field is represented as a linear combination of splines, thus enhancing the interpretability of operator networks. The variant in this case is called \textit{Operator SplineNet} (\textit{O-SNet}). The main contributions of this work are summarized as follows:
\begin{enumerate}
    \item We propose SplineNet to explicitly embed unstructured splines into neural nets, achieving watertight representations of complex geometries and seamless CAD/CAE integration in neural nets.
    \item We employ SplineNet in both solution and operator learning to enhance IGA with learning capabilities and to provide interpretability for operator networks.
    \item We apply SplineNet with Kirchhoff--Love shells to demonstrate its capability in handling real-world geometries and high-order PDEs.
\end{enumerate}

The remainder of the paper is organized as follows. Section~\ref{section 2} introduces ASUTS along with B\'ezier extraction. Section~\ref{section 3} presents the formulation of Kirchhoff--Love shells. Section~\ref{section 4} details P-SNet and O-SNet. Numerical examples are presented in Section~\ref{section 5}. Finally, Section~\ref{section 6} draws conclusions and provides suggestions for future research directions.

\section{Analysis-suitable unstructured T-splines}
\label{section 2}

Non-uniform rational B-splines (NURBS)~\cite{rogers2000introduction} are the current industrial standard for representing CAD geometries. Complex geometries are commonly represented as collections of trimmed NURBS patches. Although trimmed NURBS provide great flexibility for geometric modeling, the trimming operation generally leads to gaps and overlaps between adjacent trimmed patches, hindering the model/data exchange between CAD and CAE.

To overcome this fundamental issue, T-splines~\cite{sederberg2003t} and related variants have been developed. Among them, analysis-suitable unstructured T-splines (ASUTS)~\cite{wei2022analysis} provide desirable properties for both design and analysis, such as watertightness for complex geometries, global smoothness, linear independence, partition of unity, optimal convergence, and adaptivity. We therefore adopt ASUTS as the foundation for the seamless integration of CAD and CAE in this work. In what follows, we briefly review the basics of ASUTS. Interested readers may refer to~\cite{wei2022analysis} for details.

\begin{figure*}[tbp]
    \centering
    \begin{subfigure}[b]{0.45\textwidth}
        \centering
        \includegraphics[height=0.2\textheight]{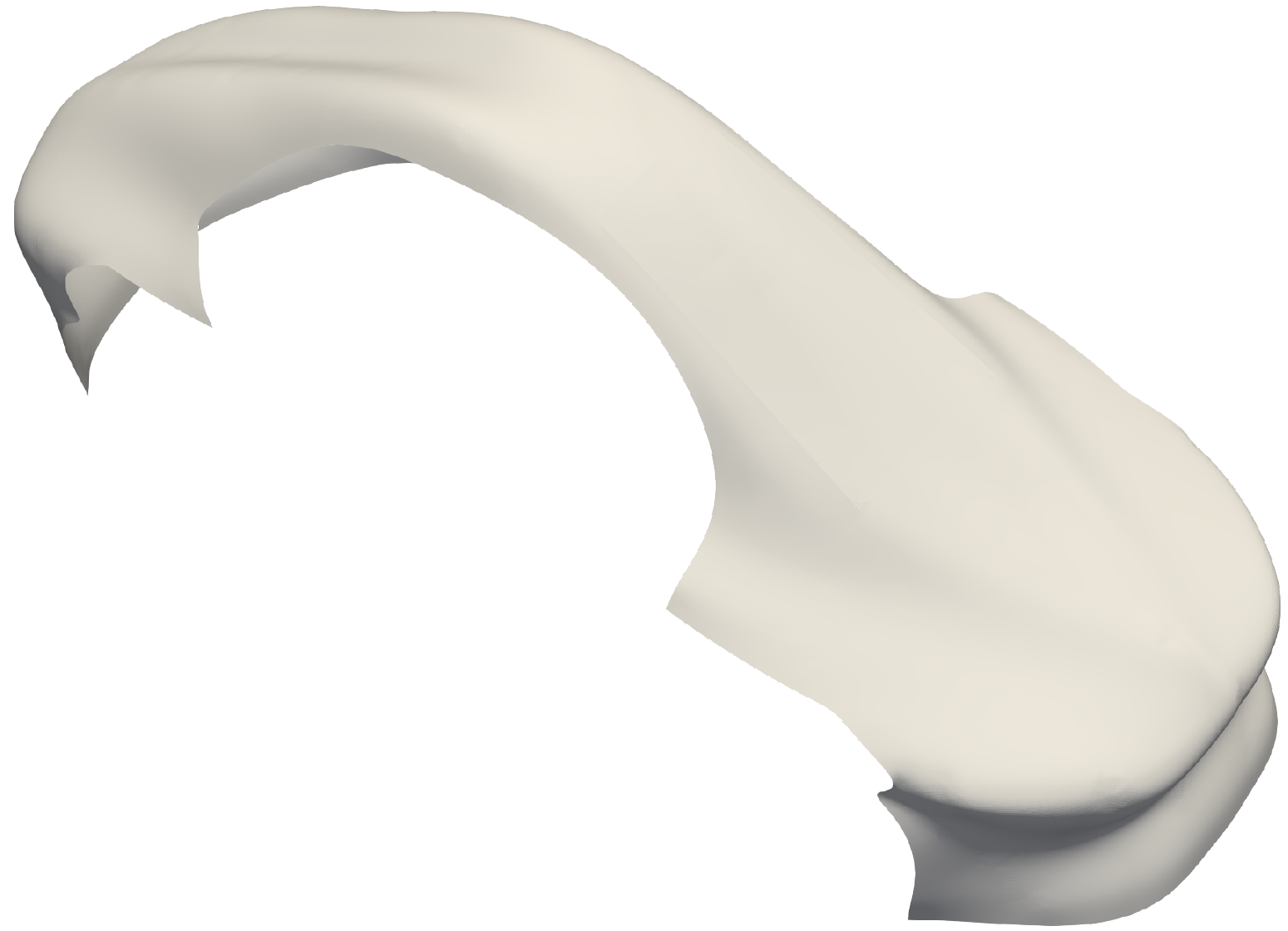}
        \caption{}
    \end{subfigure}
    \hfill
    \begin{subfigure}[b]{0.52\textwidth}
        \centering
        \includegraphics[height=0.2\textheight]{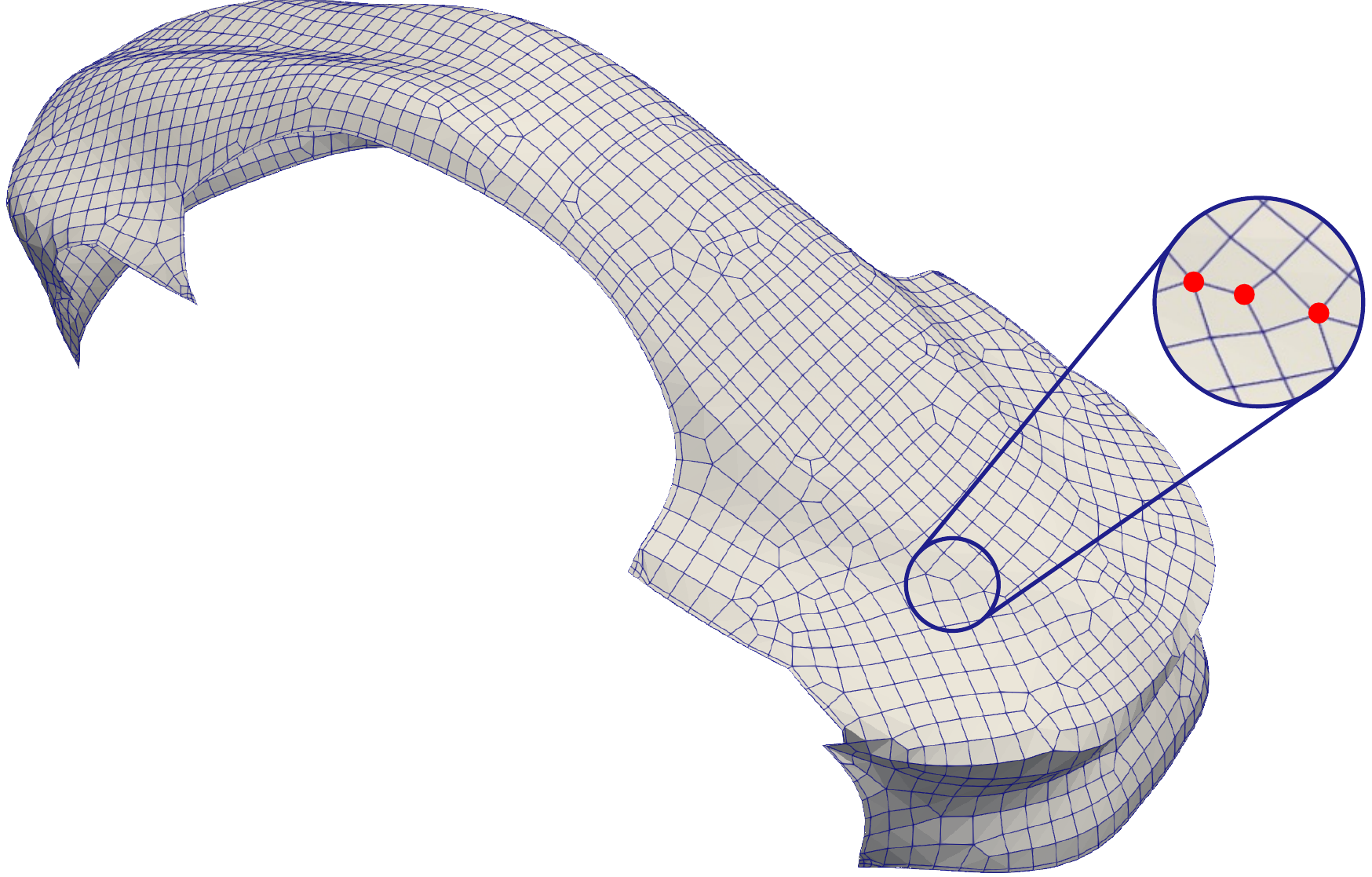}
        \caption{}
    \end{subfigure}
    \caption{Car hull model (a) and its control mesh (b) with extraordinary points (red dots).}
    \label{fig:complex_geometry}
\end{figure*}

ASUTS are built from two components: an unstructured quadrilateral control mesh (possibly with T-junctions) and the corresponding spline basis functions. The control mesh consists of control points, edges, and faces (or elements). A key to ASUTS is the treatment of extraordinary points (EPs). An EP is an interior point of valence other than four, or a boundary point of valence other than three or two, where the valence of a point is the number of edges sharing the point. EPs are inevitable for representing complex geometries; see Fig.~\ref{fig:complex_geometry}. A geometry is represented as
\begin{equation}
    \boldsymbol{r}(\xi,\eta)=
    \begin{bmatrix}
    x(\xi,\eta) \\ 
    y(\xi,\eta) \\ 
    z(\xi,\eta)
    \end{bmatrix}
    =\sum_{i=1}^n N_i(\xi,\eta)\boldsymbol{P}_i,
\end{equation}
where \(N_i(\xi,\eta)\) denotes the ASUTS basis function associated with the control point \(\boldsymbol{P}_i\), and \(n\) is the number of basis functions.

Every ASUTS basis function is a bicubic piecewise smooth polynomial. They are defined elementwise and vary with element types. According to the influence of EPs, faces in an ASUTS control mesh are mainly classified into irregular, transition, and regular elements. An irregular element is a face that contains at least one EP. Transition elements are adjacent to irregular ones. The remaining faces are regular. 
ASUTS basis functions in regular elements are simply bicubic B-splines. However, special treatments are required around EPs to ensure desired properties such as smoothness and convergence. More specifically, extra face-based points and associated splines are added in irregular elements to guarantee convergence, on top of which the D-patch method~\cite{REIF1997174} is adopted to build \(C^1\)-continuity into splines. Subsequently, the truncation mechanism~\cite{wei2018blended} is used in transition elements to make the newly added splines compatible with existing ones and also to recover partition of unity.

\begin{figure*}[tbp]
    \centering
    \begin{subfigure}[b]{0.43\textwidth}
        \centering
        \includegraphics[page=1, width=\linewidth]{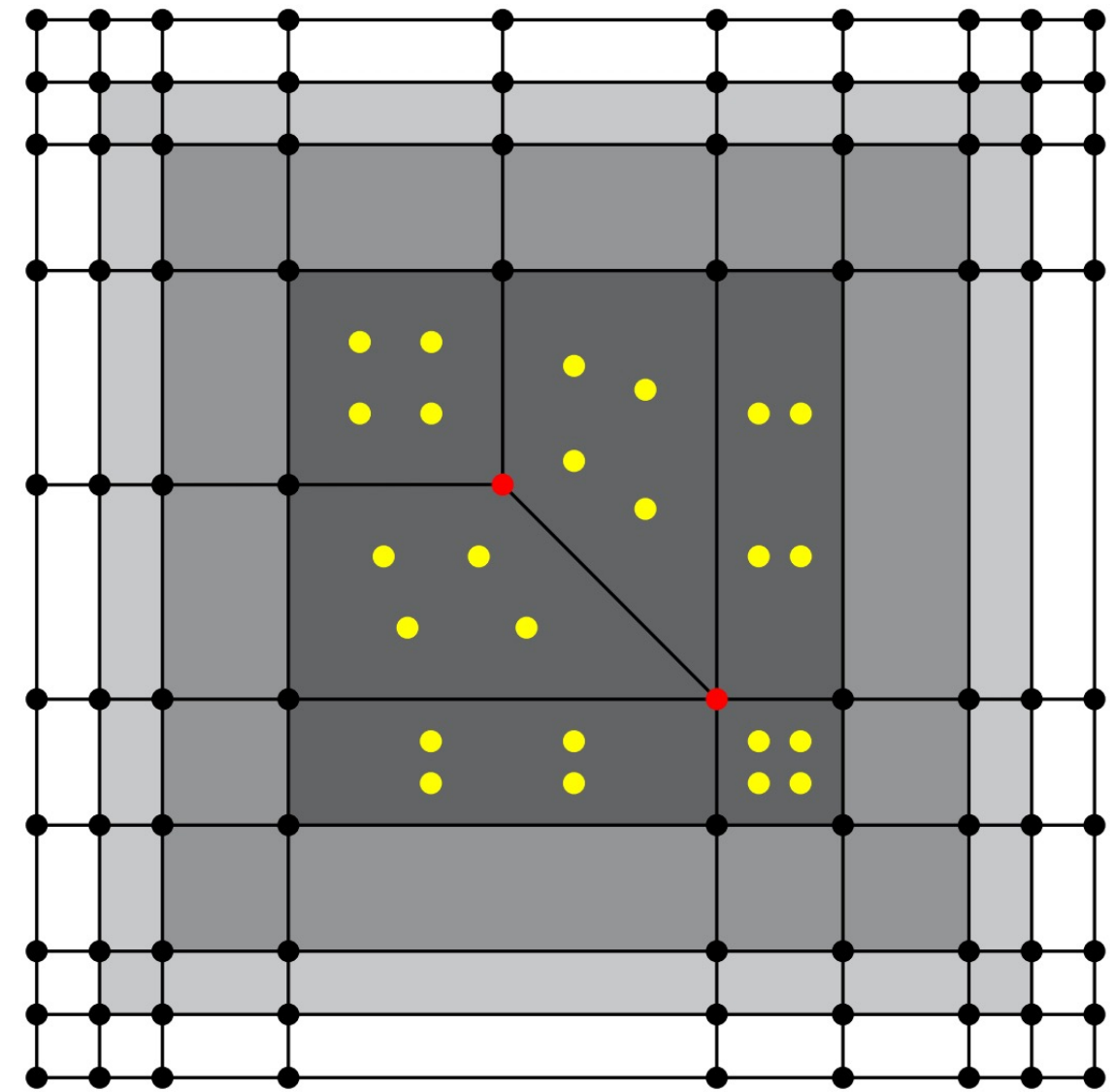}
        \caption{Control mesh}
    \end{subfigure}
    \hfill
    \begin{subfigure}[b]{0.43\textwidth}
        \centering
        \includegraphics[page=2, width=\linewidth]{figures/figure4.pdf}
        \caption{B\'ezier mesh for analysis}
    \end{subfigure}
    \caption{Example of an ASUTS control mesh and its B\'ezier mesh. (a) The control mesh contains regular control points (black dots), extraordinary points (red dots), and face-based points (yellow dots), where irregular, transition, and regular elements are shaded dark gray, gray, and light gray, respectively. (b) The corresponding B\'ezier mesh through B\'ezier extraction; note that every irregular element in (a) yields four B\'ezier elements due to the $2 \times 2$ split, which is used by ASUTS to locally increase the degrees of freedom.}
    \label{Bezier_extraction}
\end{figure*}

ASUTS basis functions are represented through B\'ezier extraction for every element~\cite{scott2011isogeometric}. On element \(e\), the local ASUTS basis functions \(\boldsymbol{N}^e(\xi,\eta)\) are expressed as a linear combination of the 16 bicubic Bernstein polynomials~\(\boldsymbol{B}(\xi,\eta)\):
\begin{equation}
    \boldsymbol{N}^e(\xi,\eta) = \boldsymbol{C}^e \boldsymbol{B}(\xi,\eta),
\label{eq:ASUTS}
\end{equation}
where \(\boldsymbol{C}^e\) is the B\'ezier extraction operator and $\boldsymbol{B}(\xi,\eta)$ is
\begin{equation}
\begin{aligned}
& [B_1(\xi,\eta),\ldots,B_{16}(\xi,\eta)]^T \\
&:= [\,b_1(\xi)b_1(\eta),\, b_2(\xi)b_1(\eta),\, \ldots,b_4(\xi)b_4(\eta)\,]^T .
\end{aligned}
\label{eq:bernstein-vector}
\end{equation}
Here $b_i(\cdot)$ is defined by
\begin{equation}
b_i(s) = \binom{3}{i-1}s^{i-1}(1-s)^{4-i},
\label{eq:cubic-bernstein}
\end{equation}
where $i=1,2,3,4$ and $s\in[0,1]$.
\(\boldsymbol{C}^e\) has dimension \(k \times 16\), where \(k\) is the number of ASUTS basis functions defined on element \(e\). 
Note that \(k=16\) in regular elements, whereas \(k\) varies in irregular and transition elements.
An example is illustrated in Fig.~\ref{Bezier_extraction}.

\begin{figure*}[tbp]
\centering 
\begin{minipage}{\linewidth}
\centering

    \begin{subfigure}[b]{0.48\linewidth} 
    \centering 
    \includegraphics[page=1,width=\linewidth]{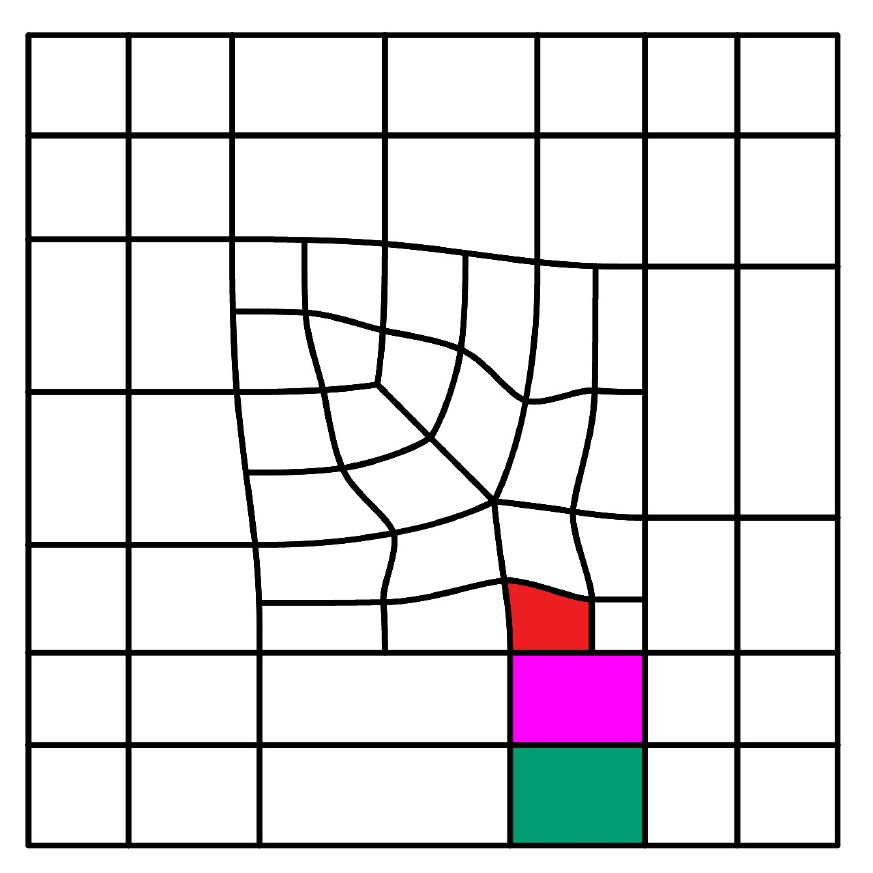} 
    \caption{} 
    \end{subfigure}
    \hfill
    \begin{subfigure}[b]{0.48\linewidth} 
    \centering 
    \includegraphics[page=4,width=\linewidth]{figures/figure5.pdf} 
    \caption{$k$ = 31} 
    \end{subfigure} 
    
    \vspace{0.5em}
    
    \begin{subfigure}[b]{0.48\linewidth} 
    \centering 
    \includegraphics[page=3,width=\linewidth]{figures/figure5.pdf} 
    \caption{$k$ = 17} 
    \end{subfigure}
    \hfill
    \begin{subfigure}[b]{0.48\linewidth} 
    \centering 
    \includegraphics[page=2,width=\linewidth]{figures/figure5.pdf} 
    \caption{$k$ = 16} 
    \end{subfigure} 

\end{minipage}
    
\caption{Supported splines on different types of elements. 
(a) Three types of elements under consideration: an irregular element (red), a transition element (magenta), and a regular element (green).
(b--d) The splines associated with the highlighted control points have support on the element of interest, i.e., the irregular, transition, and regular elements, respectively.} 
\label{fig:bzmesh and cps} 
\end{figure*}

For details, we consider three representative B\'ezier elements as shown in Fig.~\ref{fig:bzmesh and cps}.
The red, magenta, and green elements correspond to the irregular, transition, and regular B\'ezier elements, respectively. The splines associated with the highlighted control points have support on these three elements, which determine $k$. We observe that $k$ varies in different elements.

\begin{figure*}[tbp]
    \centering
    \begin{minipage}{\linewidth}
    \centering

    \begin{subfigure}[b]{0.48\linewidth}
        \centering
        \includegraphics[page=1,width=\linewidth]{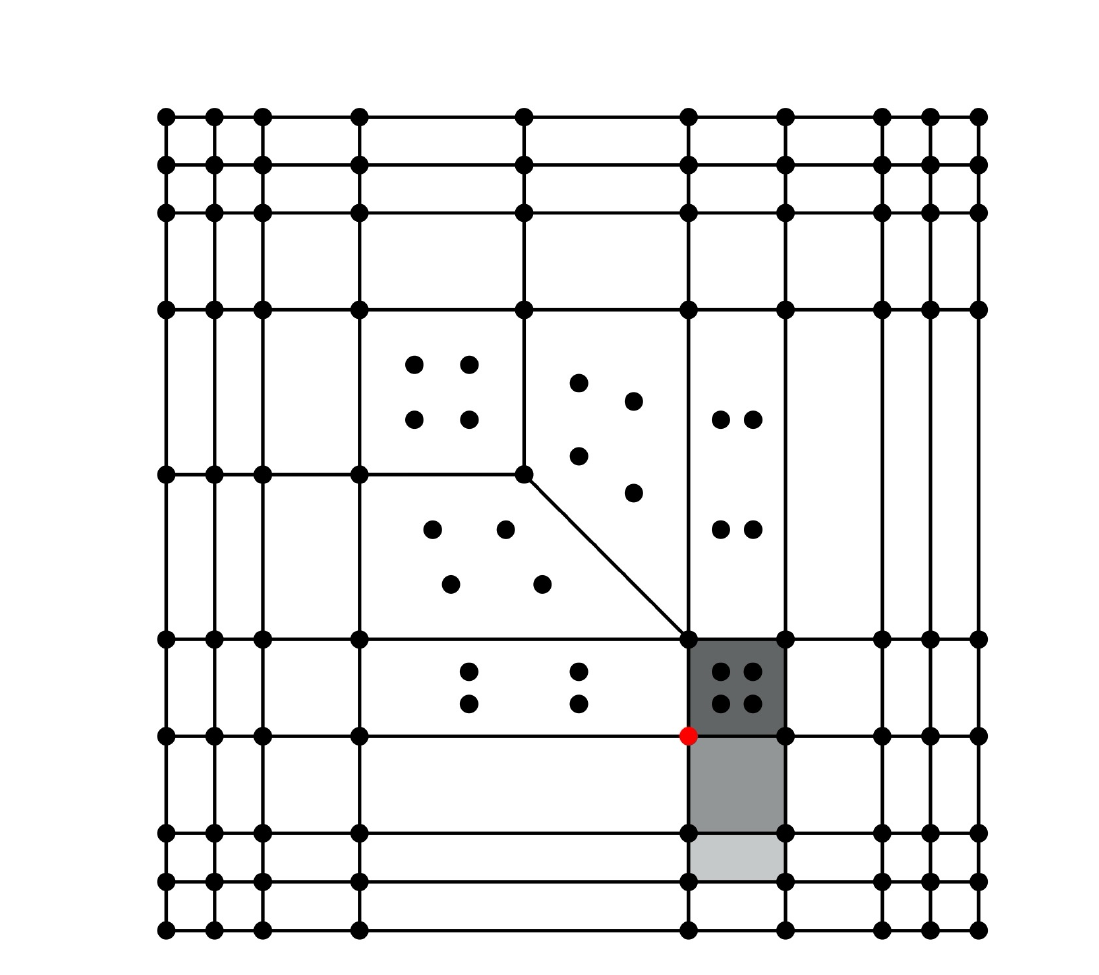}
        \caption{}
        \label{fig:cp in different elements a}
    \end{subfigure}
    \hfill
    \begin{subfigure}[b]{0.48\linewidth}
        \centering
        \includegraphics[page=2,width=\linewidth]{figures/figure6.pdf}
        \caption{}
        \label{fig:cp in different elements b}
    \end{subfigure}

    \vspace{0.5em}

    \begin{subfigure}[b]{0.48\linewidth}
        \centering
        \includegraphics[page=3,width=\linewidth]{figures/figure6.pdf}
        \caption{}
        \label{fig:cp in different elements c}
    \end{subfigure}
    \hfill
    \begin{subfigure}[b]{0.48\linewidth}
        \centering
        \includegraphics[page=4,width=\linewidth]{figures/figure6.pdf}
        \caption{}
        \label{fig:cp in different elements d}
    \end{subfigure}

    \end{minipage}

    \caption{Extraction coefficients associated with a selected spline function. 
    (a) The selected control point (marked in red). 
    (b--d) The extraction coefficients associated with the selected spline in the irregular (dark gray), transition (gray), and regular (light gray) elements, respectively. 
    Note that only the lower-left B\'ezier sub-element is shown for the irregular element.}
    \label{fig:cp in different elements}
\end{figure*}

Moreover, we pick one spline function, $N_I(\xi, \eta)$, near an EP to illustrate how it is represented elementwise; see the red dot in Fig.~\ref{fig:cp in different elements a}, which is its corresponding control point.
When restricted on element $e$, $N_I(\xi, \eta)$ is expressed as a linear combination of the 16 Bernstein polynomials,
\begin{equation}
N_i^e(\xi,\eta) = \sum_{j=1}^{16} C_{ij}^e B_j(\xi,\eta),
\label{eq:extraction_row} 
\end{equation} 
where $i$ is the local index of $I$ on element $e$, and the coefficients $C_{ij}^e$ are from $\boldsymbol{C}^e$. Their specific values corresponding to the three representative elements are shown in Fig.~\ref{fig:cp in different elements}. 
Note that the original irregular face is locally split into $2 \times 2$ B\'ezier elements, and only the lower-left one is considered here for illustration.
Such coefficients will be used to construct the proposed SplineNet.

This elementwise representation is central to the proposed method. It converts the global ASUTS into a unified elemental data structure, where each element is described by Bernstein polynomials and a precomputed extraction operator. As will be shown in Section~\ref{section 4}, this structure naturally leads to the construction of SplineNet.

\section{Kirchhoff--Love shell formulation}
\label{section 3}

The Kirchhoff--Love (KL) shell theory provides a classical model for thin-shell analysis by neglecting transverse shear deformation and requiring the shell director to remain normal to the middle surface during deformation~\cite{kiendl2009isogeometric}. The resulting weak formulation involves second-order derivatives of the displacement and therefore requires at least $C^1$-continuous basis functions. This makes ASUTS particularly suitable for KL shell analysis.

Given a thin shell $\boldsymbol{X}$ in the 3D Euclidean space, let its reference mid-surface be parameterized by
\begin{equation}
\boldsymbol{X}(\boldsymbol{\theta}) : \hat{\Omega}\subset\mathbb{R}^2 \rightarrow \mathbb{R}^3,
\quad
\boldsymbol{\theta}=(\theta^1,\theta^2),
\end{equation}
and let its deformed configuration be
\begin{equation}
\boldsymbol{x}(\boldsymbol{\theta})
=
\boldsymbol{X}(\boldsymbol{\theta})
+
\boldsymbol{u}(\boldsymbol{\theta}),
\end{equation}
where $\boldsymbol{u}$ is the mid-surface displacement. The reference and current covariant basis vectors are defined by
\begin{equation}
\boldsymbol{A}_{\alpha}=\boldsymbol{X}_{,\alpha},
\quad
\boldsymbol{a}_{\alpha}=\boldsymbol{x}_{,\alpha},
\quad \alpha=1,2,
\end{equation}
with the corresponding unit normals
\begin{equation} 
\boldsymbol{A}_{3} = \frac{\boldsymbol{A}_1\times \boldsymbol{A}_2}{\Vert \boldsymbol{A}_1\times \boldsymbol{A}_2 \Vert}, 
\quad 
\boldsymbol{a}_{3} = \frac{\boldsymbol{a}_1\times \boldsymbol{a}_2}{\Vert \boldsymbol{a}_1\times \boldsymbol{a}_2 \Vert} ,
\end{equation}
The membrane and bending strain measures are defined as
\begin{equation}
\varepsilon_{\alpha\beta}
=
\frac{1}{2}
\left(
\boldsymbol{a}_{\alpha\beta}
-
\boldsymbol{A}_{\alpha\beta}
\right),
\quad
\kappa_{\alpha\beta}
=
B_{\alpha\beta}-b_{\alpha\beta},
\label{eq:strain-measures}
\end{equation}
where
\begin{equation}
\begin{alignedat}{2}
A_{\alpha\beta}
&= \boldsymbol{A}_{\alpha} \cdot \boldsymbol{A}_{\beta},
\qquad&
a_{\alpha\beta}
&= \boldsymbol{a}_{\alpha} \cdot \boldsymbol{a}_{\beta}, \\
B_{\alpha\beta}
&= \boldsymbol{A}_{\alpha,\beta}\cdot\boldsymbol{a}_{3},
&
b_{\alpha\beta}
&= \boldsymbol{a}_{\alpha,\beta}\cdot\boldsymbol{a}_{3}.
\end{alignedat}
\label{eq:geometric derivatives}
\end{equation}

The weak form is obtained from the principle of virtual work: find $\boldsymbol{u}$ such that
\begin{equation}
\delta W_{\mathrm{int}}(\boldsymbol{u};\delta\boldsymbol{u})
=
\delta W_{\mathrm{ext}}(\delta\boldsymbol{u})
\quad
\forall \delta\boldsymbol{u},
\label{eq:virtual-work}
\end{equation}
with
\begin{equation}
\begin{aligned}
\delta W_{\mathrm{int}}
&=
\int_{A}
\left(
\delta\varepsilon_{\alpha\beta} n^{\alpha\beta}
+
\delta\kappa_{\alpha\beta} m^{\alpha\beta}
\right)
dA, \\
\delta W_{\mathrm{ext}}
&=
\int_{A}
\delta u_i f_i \, dA .
\end{aligned}
\label{eq:internal-external-work}
\end{equation}
Here $n^{\alpha\beta}$ and $m^{\alpha\beta}$ are the membrane stress resultants and bending moments, $f_i$ is the applied load, and $dA=\sqrt{|A_{\alpha\beta}|}d\theta^1 d\theta^2$. The constitutive relations follow the St.~Venant--Kirchhoff KL shell model.

The present work employs the linear KL shell~\cite{casquero2023overcoming} as the model problem. Under the small-deformation assumption, the total potential energy is written as
\begin{equation}
\Pi(\boldsymbol{u})
=
\mathcal{E}_{\mathrm{m}}(\boldsymbol{u})
+
\mathcal{E}_{\mathrm{b}}(\boldsymbol{u})
-
\mathcal{W}_{\mathrm{ext}}(\boldsymbol{u}),
\label{eq:KL shell energy}
\end{equation}
where
\begin{equation}
\begin{aligned}
\mathcal{E}_{\mathrm{m}}
&=
\frac{1}{2}
\int_A
\varepsilon_{\alpha\beta} \ n^{\alpha\beta} \ dA, \\
\mathcal{E}_{\mathrm{b}}
&=
\frac{1}{2}
\int_A
\kappa_{\alpha\beta} \ m^{\alpha\beta} \ dA, \\
\mathcal{W}_{\mathrm{ext}}
&=
\int_A
f_i \ u_i \ dA.
\end{aligned}
\label{eq:linear-kl-energy-components}
\end{equation}
This energy functional will be used in the next section to construct the physical training loss.

\section{Isogeometric deep learning method}
\label{section 4}

In this section, we introduce the details of SplineNet, which serve as the building block of the proposed method. SplineNet can be used in both solution learning and operator learning. In the former case, the Physics-informed SplineNet (P-SNet) is presented with an energy-based loss to solve single-instance problems, whereas in the latter case, SplineNet serves as the trunk net of DeepONet to enable instant predictions in a CAD/CAE-integrated manner.

\subsection{SplineNet}
\label{subsec:SplineNet}
\begin{figure*}[tbp]
    \centering
    \includegraphics[width=\linewidth]{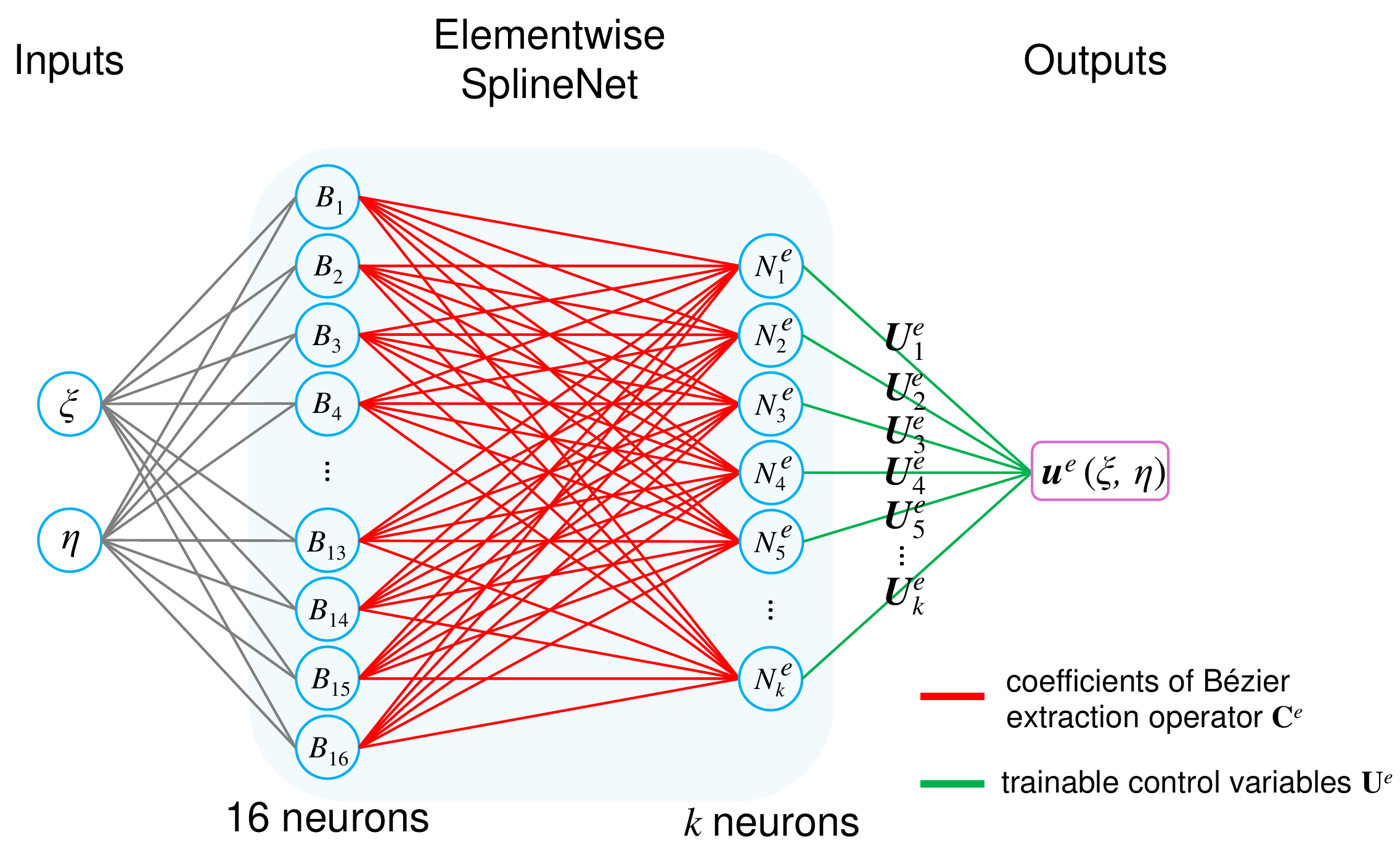}
    \caption{Architecture of the elementwise SplineNet. The input parametric coordinate \((\xi,\eta)\) is first mapped to the Bernstein polynomial layer, and then transformed into ASUTS basis functions through the B\'ezier extraction matrix \(\boldsymbol{C}^e\). The final layer combines the ASUTS basis functions with the elemental control variables to predict the displacement field on element \(e\).}
    \label{fig:SplineNet}
\end{figure*}

SplineNet embeds spline basis functions into neural networks through B\'ezier extraction, thereby enabling a CAD/CAE-integrated representation. Since both the geometry and spline basis functions are evaluated for each B\'ezier element in IGA, the elementwise SplineNet is constructed. Its architecture is illustrated in Fig.~\ref{fig:SplineNet}.

For each element, SplineNet is a fully connected network with two hidden layers. The input layer is the parametric coordinate $(\xi,\eta)$ and the output layer is the predicted solution $\boldsymbol{u}^e(\xi,\eta)$. The two hidden layers correspond to Bernstein polynomials and ASUTS basis functions, respectively.

More specifically, the first hidden layer produces Bernstein polynomials evaluated at certain parametric coordinates. It contains 16 neurons whose activation functions are the 16 bicubic Bernstein polynomials. All the weights are one, and the bias is zero. As a result, given the input parametric coordinate $(\xi, \eta)$, where $0 \leq \xi, \eta \leq 1$, the first layer simply outputs the corresponding Bernstein polynomial values \(\boldsymbol{B}(\xi,\eta)\). This layer does not involve trainable parameters.

The second hidden layer produces ASUTS basis functions as linear combinations of Bernstein polynomials. It contains \(k\) neurons, where \(k\) is the number of ASUTS basis functions defined on the element of interest (e.g., element $e$). The weights come from the extraction matrix \(\boldsymbol{C}^e\), whereas the bias is zero. Recall that an ASUTS basis function $N_i^e$ is expressed as a linear combination of Bernstein polynomials $B_j$, $N_i^e = \sum_{j=1}^{16} C_{ij}^e B_j$, where $C_{ij}^e$ is an element of $\boldsymbol{C}^e$ and it is used to define the corresponding weight in this layer. ASUTS basis functions serve as the activation functions of this layer. As a result, given the input $\boldsymbol{B}(\xi, \eta)$, this layer outputs ASUTS basis functions $\boldsymbol{N}^e(\xi, \eta)$; see Eq.~\ref{eq:ASUTS}.

The final output combines ASUTS basis functions with control variables to predict the solution at $(\xi, \eta)$. The weights are the trainable control variables $\boldsymbol{U}^e$. $\boldsymbol{U}^e$ is a $k\times d$ matrix, where $d$ (=2 or 3) is the spatial dimension. Each column of $\boldsymbol{U}^e$ is a vector of control variables along a certain direction. The bias is zero. No activation function is needed in this layer.
The output is the predicted solution, and it is obtained by 
\begin{equation}
\begin{aligned}
\boldsymbol{u}^e(\xi, \eta)
&= (\boldsymbol{U}^e)^T
   \boldsymbol{N}^e(\xi,\eta) \\
&= (\boldsymbol{U}^e)^T
   \boldsymbol{C}^e
   \boldsymbol{B}(\xi, \eta),
\end{aligned}
\end{equation}
which recovers the discretization of IGA.

\begin{remark}
    Note that, in this work, the weights of the second hidden layer are fixed as the B\'ezier extraction coefficients $\boldsymbol{C}^e$, and the bias is set to zero. With this choice, the second hidden layer exactly recovers the desired ASUTS basis functions. While these weights and biases can also be trained (which yields the \textit{r}-adaptivity), additional treatments are then required to preserve the $C^1$-continuity of ASUTS and only marginal improvement can be achieved. Therefore, we keep the weights and the bias fixed.
\end{remark}

\subsection{Physics-informed SplineNet} 
\label{subsec:P-SNet} 

\begin{figure*}[tbp]
\centering
\includegraphics[width=\linewidth]{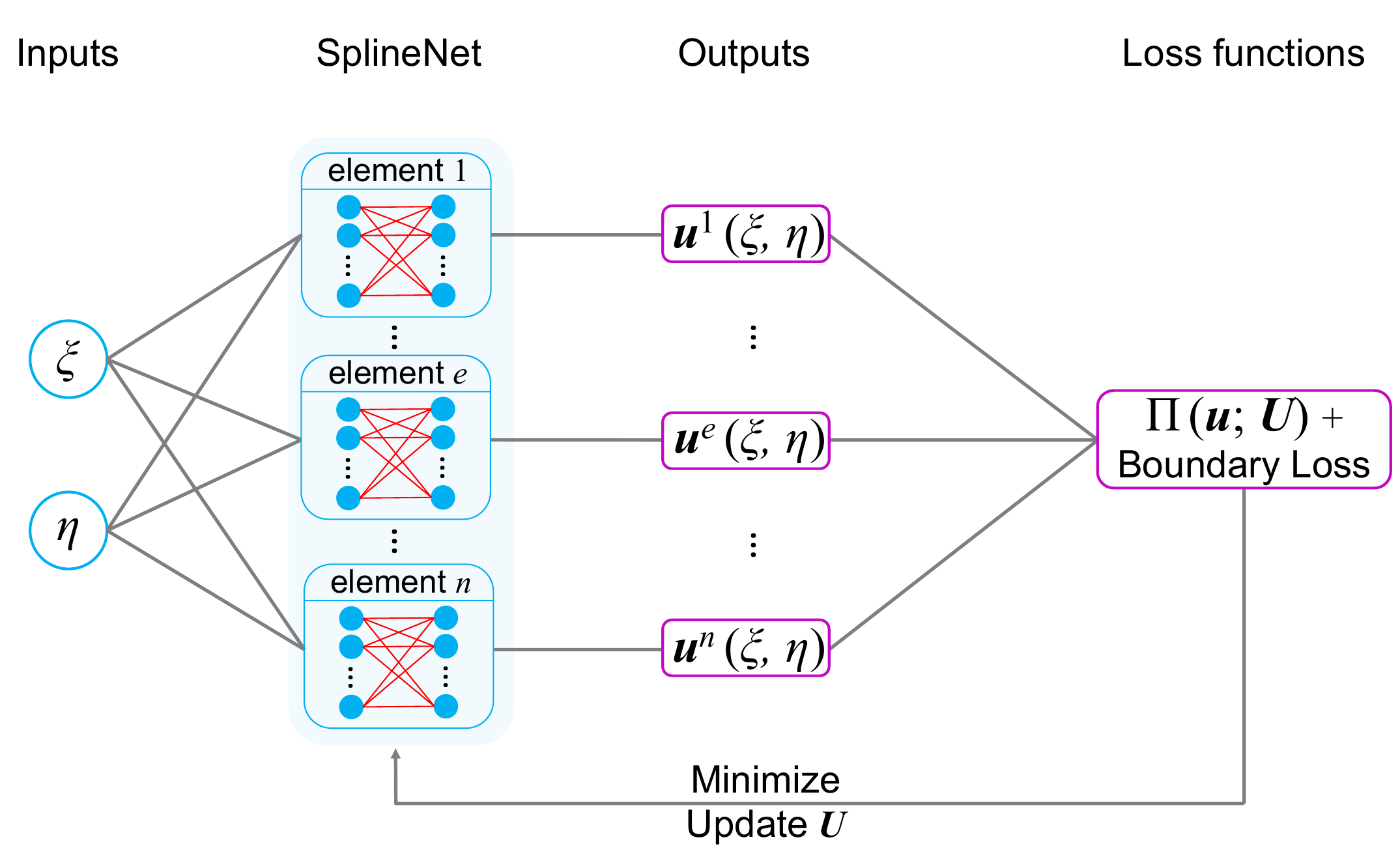}
\caption{Schematic of P-SNet.}
\label{fig:trunk net}
\end{figure*}

Physics-informed SplineNet (P-SNet) applies SplineNet to solve a single-instance problem without precomputed training data.
As shown in Fig.~\ref{fig:trunk net}, the overall SplineNet for the entire domain is obtained by stacking elementwise SplineNets for all the B\'ezier elements. In other words, each elementwise SplineNet serves as a sub-network in the overall architecture, where there are no connections among different elements, and thus sparsity is achieved to facilitate training. Note that different sub-networks generally share common control variables. In other words, a single control variable appears in several elements/sub-networks. To guarantee their uniqueness, similar to the conventional FEM/IGA, a local-to-global index map is introduced to build the correspondence between the local degrees of freedom $\boldsymbol{U}^e$ to their global ones $\boldsymbol{U}$. In practice, \(\boldsymbol{U}^e\) is directly gathered from \(\boldsymbol{U}\) by, for example, the PyTorch~\cite{paszke2019pytorch} indexing operation using the elementwise index list from ASUTS.

Each sub-network predicts the displacement $\boldsymbol{u}^e(\xi, \eta)$ restricted to its B\'ezier element, which is then used to compute the total potential energy of the KL shell. The derivatives of the geometry and displacement fields are also required to evaluate the energy.
These derivatives can be obtained in two ways. 
One option is to use automatic differentiation in PyTorch with respect to the parametric coordinates \((\xi,\eta)\). 
Alternatively, we explicitly construct derivative layers for the spline basis functions. The first hidden layer is replaced by the derivatives of the 16 Bernstein polynomials instead of the Bernstein polynomials themselves, which gives \(\boldsymbol{B}_{,\alpha}(\xi,\eta)\), where \(\alpha\in\{\xi,\eta\}\). 
The extraction matrix is then applied in the same way as in SplineNet:
\begin{equation}
\boldsymbol{N}^e_{,\alpha}(\xi,\eta)
=
\boldsymbol{C}^e
\boldsymbol{B}_{,\alpha}(\xi,\eta).
\end{equation}
Therefore, the displacement derivatives are obtained by
\begin{equation}
\boldsymbol{u}^e_{,\alpha}(\xi,\eta)
=
(\boldsymbol{U}^e)^T
\boldsymbol{N}^e_{,\alpha}(\xi,\eta).
\end{equation}
Higher-order derivatives are computed in the same manner by replacing the first hidden layer with the corresponding higher-order derivatives of the Bernstein polynomials. 
The same procedure is used for the geometric mapping by replacing the displacement control variables \(\boldsymbol{U}^e\) with the control point coordinates. 
Thus, the geometric derivatives \(\boldsymbol{a}_{\alpha}\) and \(\boldsymbol{a}_{\alpha,\beta}\) in Eq.~\ref{eq:geometric derivatives} are obtained with the displacement derivatives.
The Jacobian of the geometry mapping is further constructed from these geometric derivatives.
Moreover, the derivatives of \(\boldsymbol{u}^e\) with respect to the physical coordinates $(x, y, z)$ can be obtained by the chain rule if needed.

During training, the global control variables $\boldsymbol{U}$ are updated by minimizing the loss term that combines the total potential energy and the Dirichlet boundary condition:
\begin{equation}
    \mathcal{L}(\boldsymbol{U}) = 
    \Pi(\boldsymbol{u};\boldsymbol{U}) + 
    \frac{1}{M}
    \sum_{b=1}^{M}
    \left\|
    \boldsymbol{\hat{u}}(\xi_b, \eta_b)
    -
    \boldsymbol{u}(\xi_b, \eta_b)
    \right\|^2,
\end{equation}
where the energy $\Pi(\boldsymbol{u};\boldsymbol{U})$ is defined in Eq.~\ref{eq:KL shell energy}, $M$ is the number of predefined boundary points, and $\boldsymbol{\hat{u}}(\xi_b, \eta_b)$ and $\boldsymbol{u}(\xi_b, \eta_b)$ are given and predicted displacements of the chosen boundary points, respectively.
Usually, these boundary points are equally spaced in the parametric domain. In this work, the prescribed boundary displacements are imposed directly on the corresponding boundary control variables.
To accurately evaluate $\Pi$, the input parametric coordinates are chosen to be the Gaussian points during training. This way, P-SNet features a spline representation and enforces the shell governing physics through the energy formulation.

\subsection{Operator Spline Net}
\label{subsec:O-SNet}

Operator SplineNet (O-SNet) extends SplineNet to learn mappings from input functions (e.g., loadings) to shell structural responses. The architecture is developed based on DeepONet~\cite{lu2021learning}, which can be used to provide instant predictions for new input without retraining.

\begin{figure}[htbp]
\centering
\includegraphics[width=\linewidth]{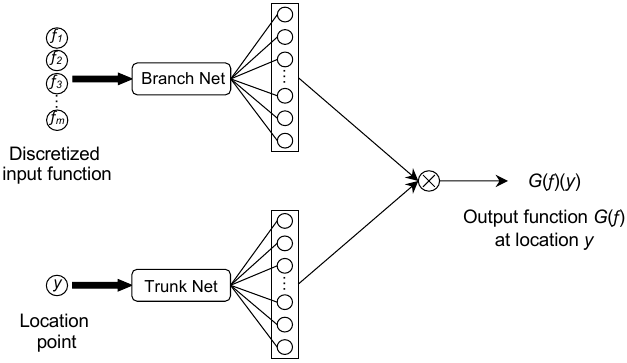}
\caption{Architecture of DeepONet. The branch net encodes the discretized input function evaluated at sensor points, while the trunk net encodes the query location. The two latent vectors are coupled to approximate the target operator and predict the output function value at the query location.}
\label{fig:deeponet}
\end{figure}

A standard DeepONet contains two networks: a branch net and a trunk net. The branch net is used to encode the input function, while the trunk net is used to encode the query location where the output function is evaluated. The outputs of them are then coupled to approximate the target operator. Different kinds of neural architectures can be used in both networks, such as multi-layer perceptrons (MLP) and convolutional neural networks.
In practice, the input function is represented by discrete values at a group of sensor points. For function \(f\), these values can be written as
$
\boldsymbol{f}
=
\left[
f(\boldsymbol{x}_1),
f(\boldsymbol{x}_2),
\ldots,
f(\boldsymbol{x}_m)
\right]^T,
$
where $\boldsymbol{x}_i$ is a sensor point and $m$ is the number of sensor points. The branch net takes \(\boldsymbol{f}\) as input and outputs a latent vector. Meanwhile, the trunk net takes the query locations $\boldsymbol{y}$ as input and outputs another latent vector. The two latent vectors are coupled through an inner product to predict the output function value at \(\boldsymbol{y}\), i.e., $G(\boldsymbol{f})(\boldsymbol{y})$, where $G$ is the mapping/operator from the input function to the output function. A schematic of DeepONet is shown in Fig.~\ref{fig:deeponet}.

\begin{figure*}[tbp]
\centering
\includegraphics[width=\linewidth]{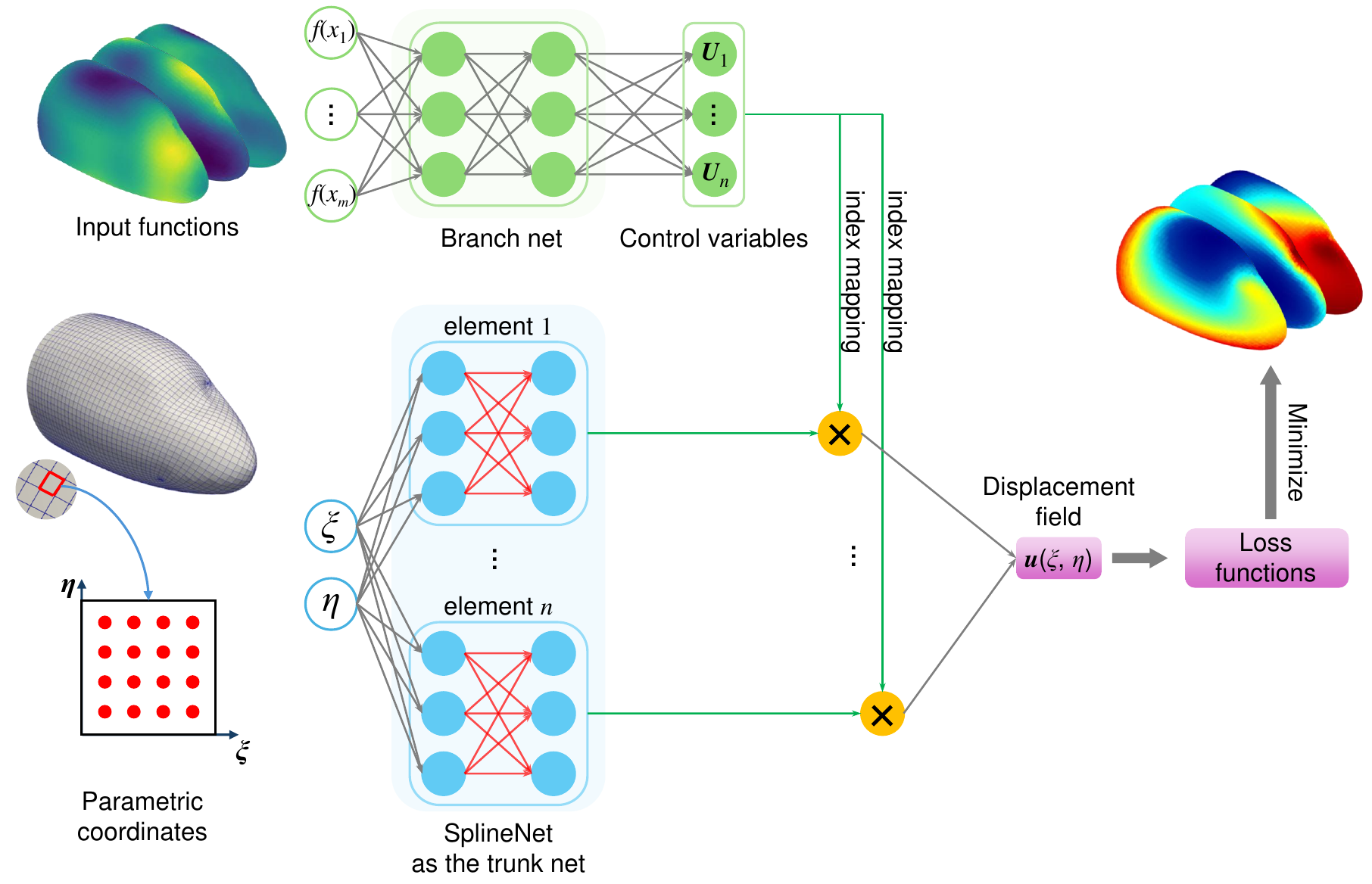}
\caption{Architecture of O-SNet. The branch net maps discretized loading functions to global control variables $\boldsymbol{U}$, while SplineNet serves as the trunk net to provide the ASUTS basis functions. Their combination yields the displacement field in the same spline space as IGA, enabling operator learning in a CAD/CAE-integrated representation.}
\label{fig:O-SNet}
\end{figure*}

In O-SNet, this structure is adapted to the isogeometric setting mainly through the trunk net, as shown in Fig.~\ref{fig:O-SNet}.
The trunk net of O-SNet features SplineNet rather than other options (e.g., MLP). Ultimately, the trunk net of O-SNet provides the ASUTS basis functions $\boldsymbol{N}(\xi, \eta)$. Doing so can not only explicitly embed high-fidelity geometric representations in a learning-based method, but also enhance IGA with learning capabilities. Instead of letting a neural net learn the basis of the output function, as is done in DeepONet, O-SNet introduces a prior to fulfill the same task. As a result, O-SNet provides predictions in the same discretized form as the conventional IGA, and thus enhances the interpretability of DeepONet. Moreover, the performance of O-SNet is independent of query point locations. Indeed, the trunk net of O-SNet remains the same no matter where the query points are chosen.

The branch net of O-SNet is simply MLP. It remains mostly the same as DeepONet except the fact that it encodes the input function $\boldsymbol{f}$ into control variables $\boldsymbol{U}(\theta)$, where $\theta$ denotes the trainable parameters of the branch net.
When combining with the output of the trunk net, O-SNet yields the desired prediction,
$\boldsymbol{u}(\xi, \eta; \theta)=\mathbf{U}(\theta)^T \boldsymbol{N}(\xi, \eta)$. 
This way, O-SNet preserves the overall structure of DeepONet while keeping the output field in the same spline space for CAD/CAE integration.

The operator learning process in O-SNet is carried out in a data-driven manner. A set of labeled data $\{\boldsymbol{f}^{(i)}, \boldsymbol{\hat{u}}^{(i)} \}^N_{i=1}$ is given, where $\boldsymbol{f}^{(i)}$ denotes a discretized input function evaluated at sensor points, and $\boldsymbol{\hat{u}}^{(i)}$ is the corresponding displacement field obtained from an FEM/IGA solver. The predicted displacement field corresponding to $\boldsymbol{f}^{(i)}$ is denoted by $\boldsymbol{u}^{(i)}$.

The loss function consists of two terms: the mean squared error (MSE) of the displacement at sampling points $(\xi, \eta)$ and the boundary conditions,
\begin{equation} 
\begin{aligned} 
\mathcal{L}(\theta) &= \mathcal{L}_{\mathrm{data}} + \mathcal{L}_{\mathrm{bc}} \\ 
&= \frac{1}{NK} \sum_{i=1}^{N} \sum_{j=1}^{K} \left\| \boldsymbol{\hat{u}}^{(i)}(\xi_{j}, \eta_{j}) - \boldsymbol{u}^{(i)}(\xi_{j}, \eta_{j};\theta) \right\|^2 \\ 
&+ \frac{1}{NM} \sum_{i=1}^{N} \sum_{b=1}^{M} \left\| \boldsymbol{\hat{u}}^{(i)}(\xi_{b}, \eta_{b}) - \boldsymbol{u}^{(i)}(\xi_{b}, \eta_{b};\theta) \right\|^2, 
\end{aligned} 
\label{eq:loss_OSNet} 
\end{equation}
where $K$ denotes the number of sampling points for the output, and $M$ denotes the number of boundary points. In this work, boundary conditions are enforced through control variables.

\section{Numerical examples}
\label{section 5}

In this section, numerical examples are presented to assess the proposed method. The first example verifies P-SNet as a data-free energy-based solver using the classical Scordelis--Lo roof benchmark. The remaining examples evaluate O-SNet for operator learning on three shell structures with increasing geometric complexity: a roof with extraordinary points, a B-pillar, and a plane nose. These tests examine the accuracy and generalization capability of the proposed method.

\subsection{Solution learning with P-SNet}

\begin{figure*}[tbp]
    \centering

    \begin{subfigure}[b]{0.48\linewidth}
        \centering
        \includegraphics[page=1,width=\linewidth]{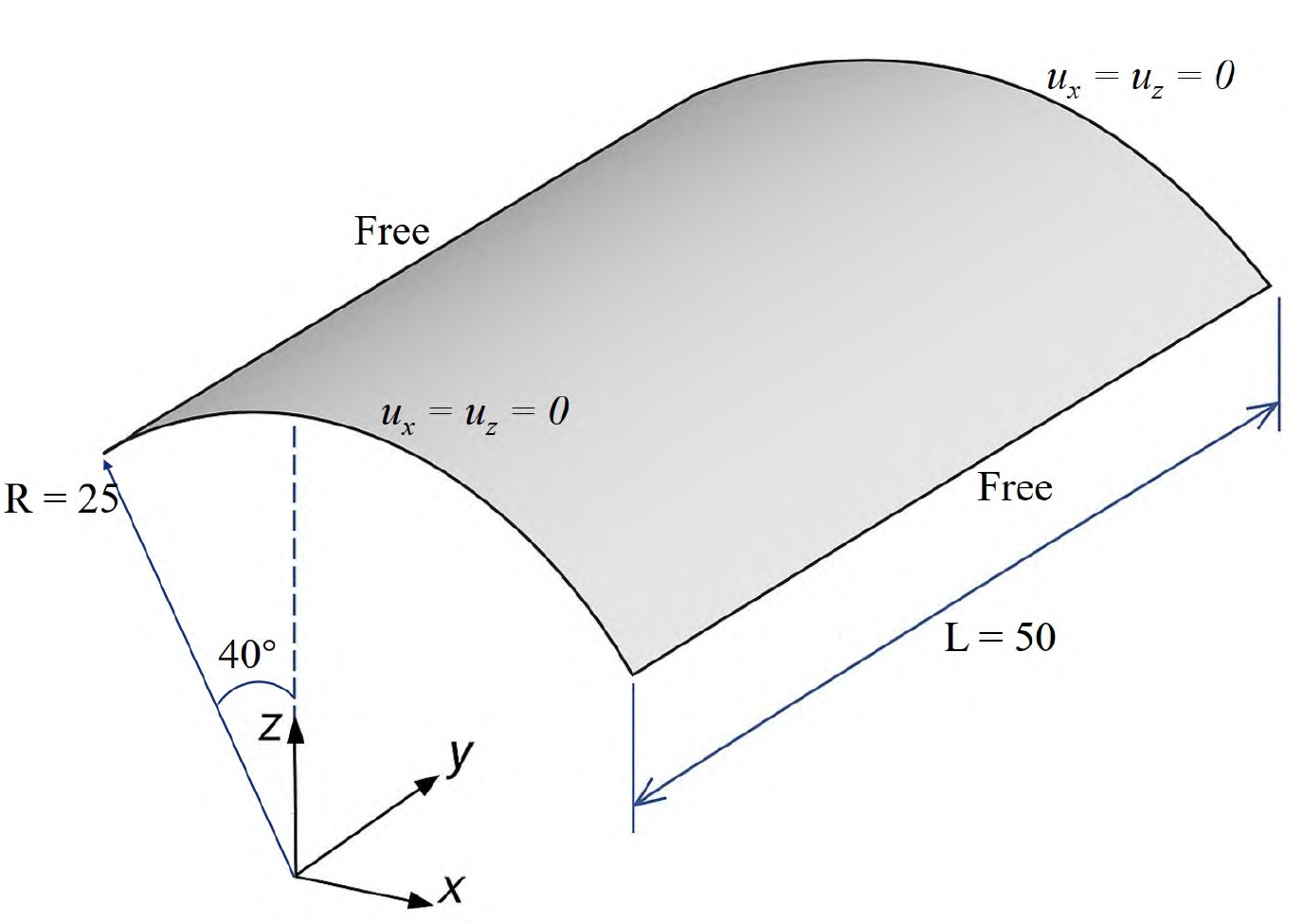}
        \caption{Geometry and boundary conditions}
        \label{fig:scordelis_bezier_mesh a}
    \end{subfigure}
    \hfill
    \begin{subfigure}[b]{0.48\linewidth}
        \centering
        \includegraphics[page=2,width=\linewidth]{figures/figure11.pdf}
        \caption{Mesh 0}
        \label{fig:scordelis_bezier_mesh b}
    \end{subfigure}

    \vspace{0.5em}

    \begin{subfigure}[b]{0.48\linewidth}
        \centering
        \includegraphics[page=3,width=\linewidth]{figures/figure11.pdf}
        \caption{Mesh 1}
        \label{fig:scordelis_bezier_mesh c}
    \end{subfigure}
    \hfill
    \begin{subfigure}[b]{0.48\linewidth}
        \centering
        \includegraphics[page=4,width=\linewidth]{figures/figure11.pdf}
        \caption{Mesh 2}
        \label{fig:scordelis_bezier_mesh d}
    \end{subfigure}

    \caption{
    Scordelis--Lo roof and its unstructured B\'ezier meshes.
    (a) Problem settings: radius \(R=25\,\mathrm{m}\), length \(L=50\,\mathrm{m}\), thickness \(t=0.25\,\mathrm{m}\), uniform gravity load \(g=90\,\mathrm{N/m^2}\), Young's modulus \(E=432\,\mathrm{MPa}\), and Poisson's ratio \(\nu=0.0\).
    (b) Initial mesh with 48 elements and 124 control points.
    (c) Refined mesh with 120 elements and 432 control points.
    (d) Refined mesh with 408 elements and 1624 control points.
    }
    \label{fig:scordelis_bezier_mesh}
\end{figure*}

To verify the performance of P-SNet, we start with the Scordelis--Lo roof problem, which has been widely used as a benchmark for shells. The roof is subjected to gravity and supported at the two curved edges.
Three consecutively refined meshes are considered where EPs are intentionally introduced to test ASUTS in the context of SplineNet; see Fig.~\ref{fig:scordelis_bezier_mesh}.

\begin{figure*}[tbp]
    \centering
    \begin{minipage}{\linewidth}
    \centering
    
    \includegraphics[width=\linewidth]{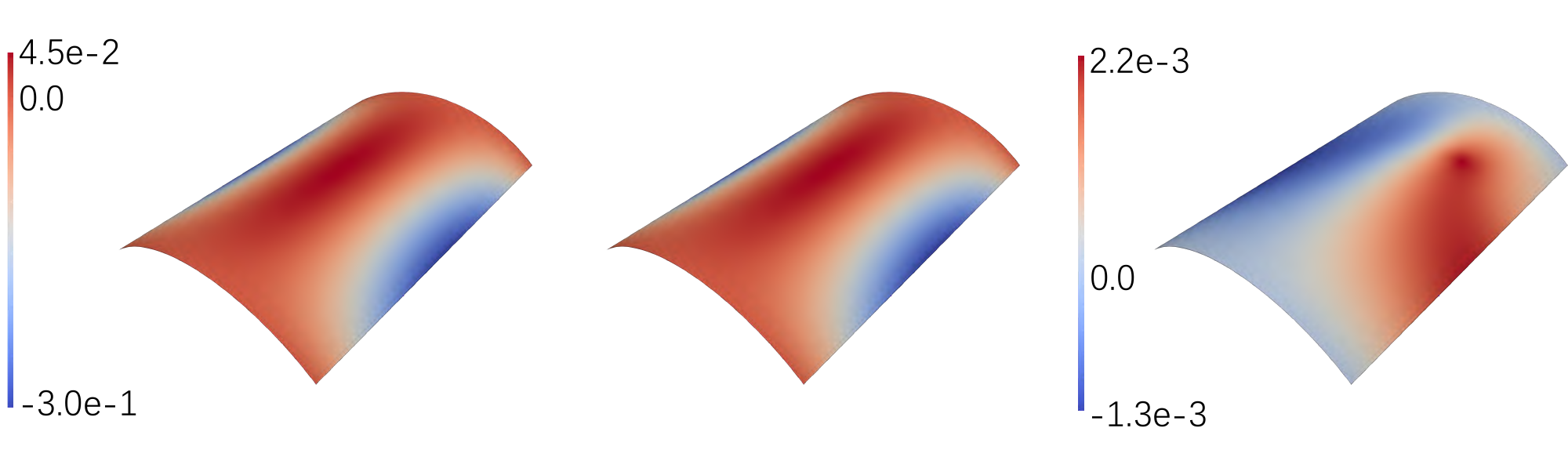}
    
    \vspace{0.3em}
    
    \makebox[\linewidth][l]{%
        \hspace*{0.166\linewidth}\makebox[0pt][c]{(a) IGA result}%
        \hspace*{0.333\linewidth}\makebox[0pt][c]{(b) Prediction by P-SNet}%
        \hspace*{0.333\linewidth}\makebox[0pt][c]{(c) Error}%
    }
    
    \end{minipage}
    \vspace{0.5em}
    \caption{Displacement ($z$ component) comparison for the Scordelis--Lo roof on Mesh 2.}
    \label{fig:scordelis_roof_tests}
\end{figure*}

To keep the presentation concise, Fig.~\ref{fig:scordelis_roof_tests} reports the displacement field comparison on the finest mesh, whereas the free-edge midpoint displacements and errors for all meshes are summarized in Table~\ref{tab:psnet_scordelis_midpoint}. 
For all the three meshes, the relative errors of the free-edge midpoint displacement remain below \(1\%\). 
In particular, on the finest mesh, P-SNet achieves a relative error of \(0.679\%\), and the predicted displacement field closely matches the IGA reference solution.
These results demonstrate that P-SNet can accurately predict the solution of the Scordelis--Lo roof with unstructured ASUTS meshes, validating its effectiveness as an energy-based solver for single-instance shell problems.

\begin{table*}[tbp]
\caption{Free-edge midpoint displacement comparison.}
\centering
\begin{tabular*}{\textwidth}{@{\extracolsep{\fill}}ccccc@{}}
\hline
Mesh & \(u_\mathrm{mid}\) (IGA) & \(u_\mathrm{mid}\) (P-SNet) & \(|\Delta u_\mathrm{mid}|\) & Relative error (\%) \\ 
\hline
Mesh 0 & -0.25716 & -0.25723 & 0.00007 & 0.027 \\
Mesh 1 & -0.29781 & -0.29780 & 0.00001 & 0.003 \\
Mesh 2 & -0.30038 & -0.30242 & 0.00204 & 0.679 \\
\hline
\end{tabular*}
\label{tab:psnet_scordelis_midpoint}
\end{table*}

\subsection{Operator learning with O-SNet}
In this section, we investigate operator learning using O-SNet. Unlike P-SNet, which solves one fixed boundary value problem at a time, O-SNet learns the mapping from loading functions to structural responses. Several shell structures with EPs are studied to evaluate the performance of O-SNet.

\subsubsection{Data generation and parameter settings}
In this work, the loading functions are generated by the mean-zero Gaussian random field (GRF):
\begin{equation}
    f(\boldsymbol{x}) \sim \gamma {G}\big(0, k(\boldsymbol{x}, \boldsymbol{x}')\big), 
\end{equation}
where $ k(\boldsymbol{x}, \boldsymbol{x}') = \exp\!\left(-\frac{\|\boldsymbol{x}-\boldsymbol{x}'\|^2}{2 l^2}\right)$ is the Gaussian kernel, $l$ denotes the length scale controlling the smoothness of loading functions, and $\gamma$ is the amplitude factor that amplifies the magnitude of loading functions. In this work, the length scale is set to $l=0.2$, and the amplitude factor varies in different problems. The loading functions are sampled at the 16 Gaussian quadrature points of each B\'ezier element, which are also used as the inputs to SplineNet.

For the linear Kirchhoff--Love shell tests (roof and B-pillar), the parameters are listed in Table~\ref{material property}. The structural displacement data are obtained from the in-house IGA solver~\cite{sheng2025isogeometric} based on the ASUTS. The O-SNet model is implemented in PyTorch~\cite{paszke2019pytorch} and trained on an NVIDIA RTX 6000 Ada Generation GPU.

\begin{table}[tbp]
\caption{Parameters of KL shells.}
\centering
\begin{tabular*}{\linewidth}{@{\extracolsep{\fill}}ccc@{}}
\hline
Parameters & Meaning & Value \\
\hline
$E$   & Young's modulus  & $2 \times 10^{11}$  \\
$\nu$ & Poisson's ratio & 0.3                 \\
$t$   & thickness        & 0.02                \\
\hline
\end{tabular*}
\label{material property}
\end{table}

\subsubsection{Roof}

\begin{figure}[tbp]
    \centering

    
    \begin{subfigure}{\linewidth}
        \centering
        \includegraphics[page=1, width=\linewidth]{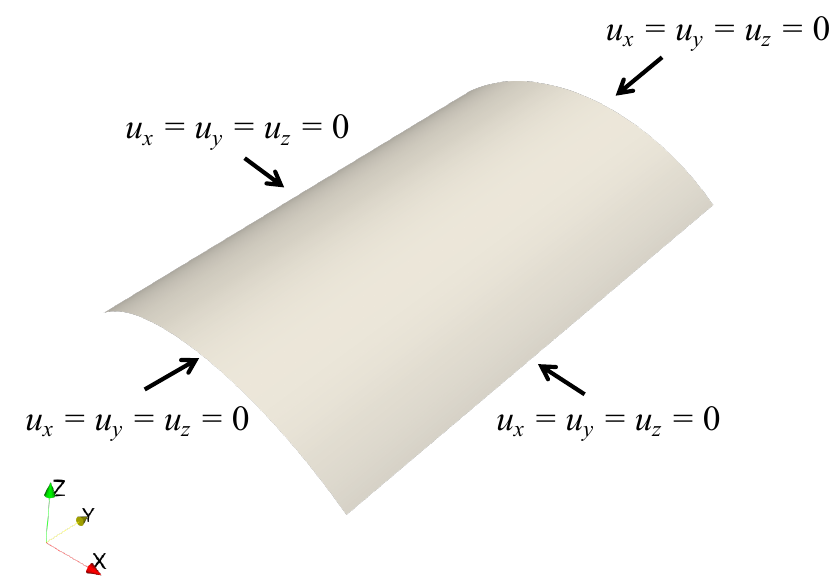}
        \caption{}
        \label{fig:roof_model}
    \end{subfigure}
    \hfill
    \begin{subfigure}{\linewidth}
        \centering
        \includegraphics[page=2, width=\linewidth]{figures/figure13.pdf}
        \caption{}
        \label{fig:roof_bzmesh}
    \end{subfigure}

    \caption{The roof problem for operator learning. (a) Problem setting with four fully clamped edges, subjected to $z$-direction loading over the entire surface, and (b) the corresponding B\'ezier mesh consisting of 120 elements and 432 control points.}
    \label{fig:roof_case}
\end{figure}

We begin with a simple roof problem. 
The geometric parameters are identical to those of the Scordelis--Lo roof, except that a different thickness is used and all four edges are fully clamped, as shown in Fig.~\ref{fig:roof_case}. The roof is subjected to distributed surface loadings in the $z$-direction, generated using GRF with an amplitude factor of $1.5 \times 10^8$. The roof is represented by ASUTS, where two EPs are intentionally introduced.

A total of 1250 loading functions are generated, with 1000 used for training and the remaining 250 for testing. The branch net is implemented as an MLP with two hidden layers. We compare three network widths with 100, 300, and 500 neurons per hidden layer.

\begin{figure}[tbp]
    \centering

    \begin{subfigure}[t]{0.8\linewidth}
        \centering
        \includegraphics[page=1, width=\linewidth]{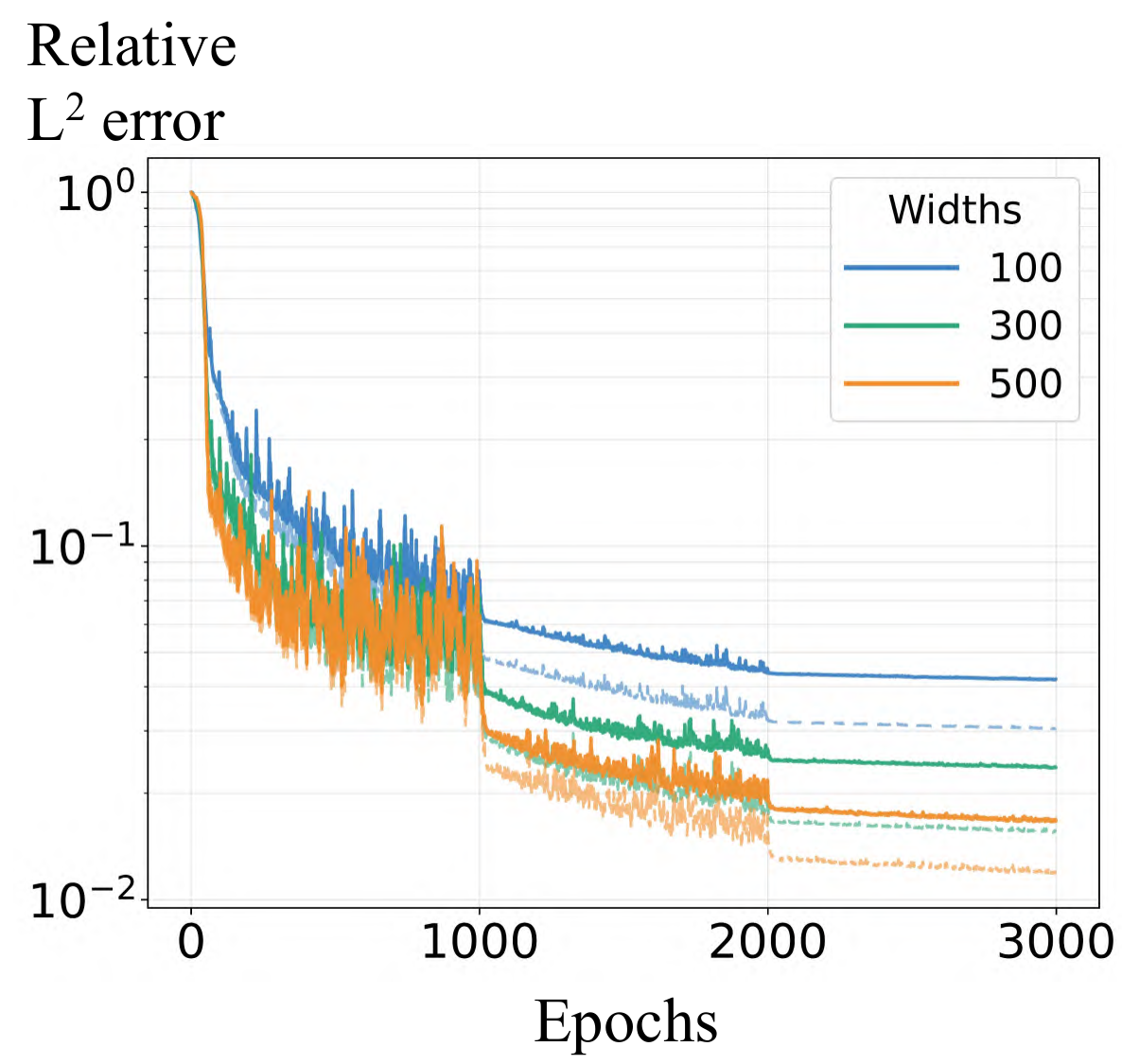}
        \caption{}
        \label{fig:roof L2 a}
    \end{subfigure}
    \hfill
    \begin{subfigure}[t]{0.8\linewidth}
        \centering
        \includegraphics[page=2, width=\linewidth]{figures/figure14.pdf}
        \caption{}
        \label{fig:roof L2 b}
    \end{subfigure}

    \caption{Training curve and network-width comparison for the roof problem. 
    (a) Mean relative \(L^2\) error versus epochs, where solid and dashed lines denote test and training errors, respectively (with learning rates \(10^{-3}\), \(10^{-4}\), and \(10^{-5}\), each used for 1000 epochs).
    (b) Minimum relative \(L^2\) errors for different network widths, evaluated at Gaussian quadrature points (GPs) and element corners (ECs).}
    \label{fig:roof L2}
\end{figure}

All models are trained for 3000 epochs using the Adam optimizer~\cite{kingma2014adam} with the staged learning rate schedule shown in Fig.~\ref{fig:roof L2}. 
The mean relative $L^2$ error of the $z$-displacement is evaluated on both training and test sets at Gaussian quadrature points (GPs) and element corners (ECs). Only the GP values are used for training, whereas the EC errors are reported to assess the predicted control variables. Fig.~\ref{fig:roof L2} summarizes the training process and the best mean errors obtained with different network widths. 

As shown in Fig.~\ref{fig:roof L2 a}, the training and test errors decrease consistently as the number of epochs increases.
The small gap between the training and test curves suggests that no obvious overfitting/underfitting occurs. 
Fig.~\ref{fig:roof L2 b} further compares different network widths. 
Increasing the width generally improves the prediction accuracy at both GPs and ECs, and the model with 500 neurons per hidden layer gives the lowest errors. 
Therefore, this model is selected for the following comparisons. 
The EC errors are slightly larger than the GP errors because the EC values are not directly used in the loss function. 
Table~\ref{tab:roof_error} reports the average $L^2$ errors.
\begin{figure*}[p]
    \centering
    \begin{minipage}{\linewidth}
    \centering
    \includegraphics[width=\linewidth]{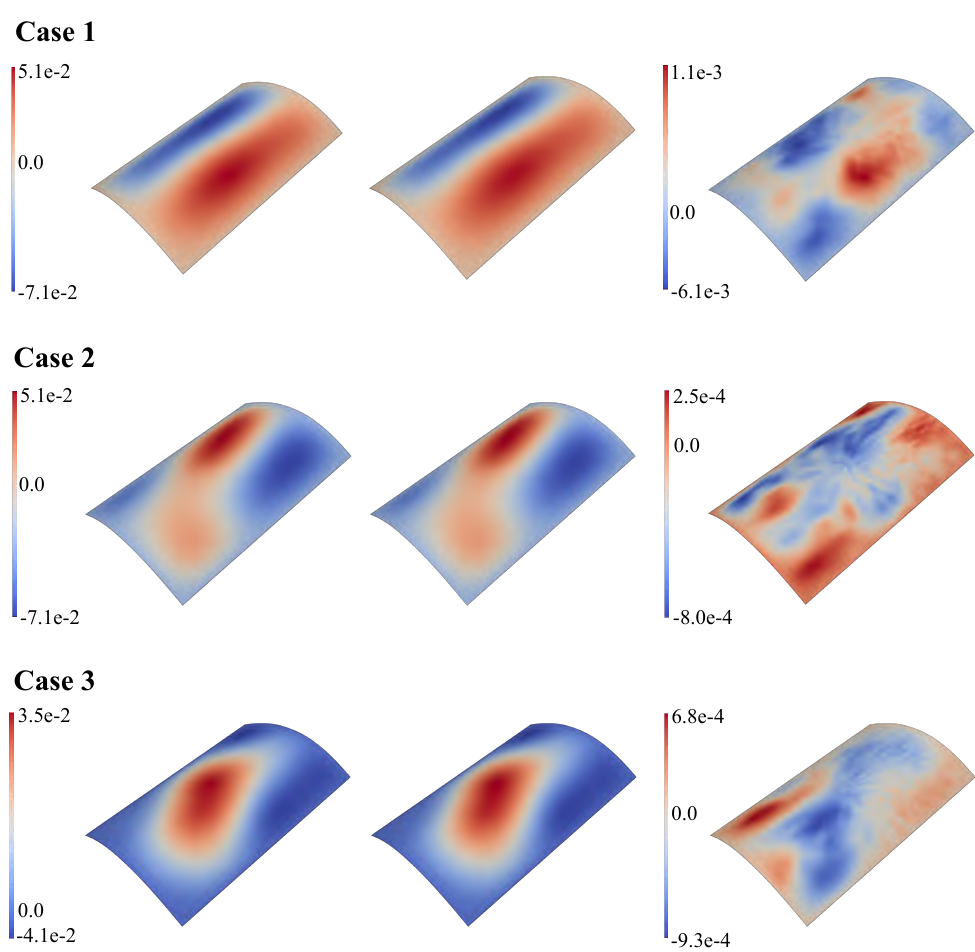}
    
    \vspace{0.3em}
    
    \makebox[\linewidth][l]{%
        \hspace*{0.20\linewidth}\makebox[0pt][c]{(a) IGA result}%
        \hspace*{0.31\linewidth}\makebox[0pt][c]{(b) Prediction by O-SNet}%
        \hspace*{0.34\linewidth}\makebox[0pt][c]{(c) Error}%
    }

    \end{minipage}

    \vspace{0.3em}
    
    \caption{Comparison of $z$-displacement fields for three representative test cases of the roof problem.}
    \label{fig:roof cases}
\end{figure*}

To further illustrate the performance, three representative cases are selected from the test dataset corresponding to the first (Case 1), middle (Case 2), and last sample (Case 3).
The predicted \(z\)-displacement fields are compared with the IGA reference solutions in Fig.~\ref{fig:roof cases}. 
The relative $L^2$ errors at GPs and ECs are $1.27\%$ and $1.46\%$ for Case 1, $1.72\%$ and $2.20\%$ for Case 2, and $1.90\%$ and $2.95\%$ for Case 3, respectively. 
The predictions closely match the reference solutions. The larger EC errors are a result of the fact that EC values are not directly enforced during training.

\begin{table}[tbp]
\caption{Average relative $L^2$ errors of the $z$-direction displacement in the roof problem with 500 neurons.}
\centering
\begin{tabular*}{\linewidth}{@{\extracolsep{\fill}}ccc@{}}
\hline
Error type & Training set & Test set \\
\hline
GPs & 1.19\% & 1.66\% \\
ECs & 2.21\% & 2.56\% \\
\hline
\end{tabular*}
\label{tab:roof_error}
\end{table}

\FloatBarrier
\subsubsection{B-pillar}

\begin{figure}[tbp] 
\centering 
\includegraphics[width=\linewidth]{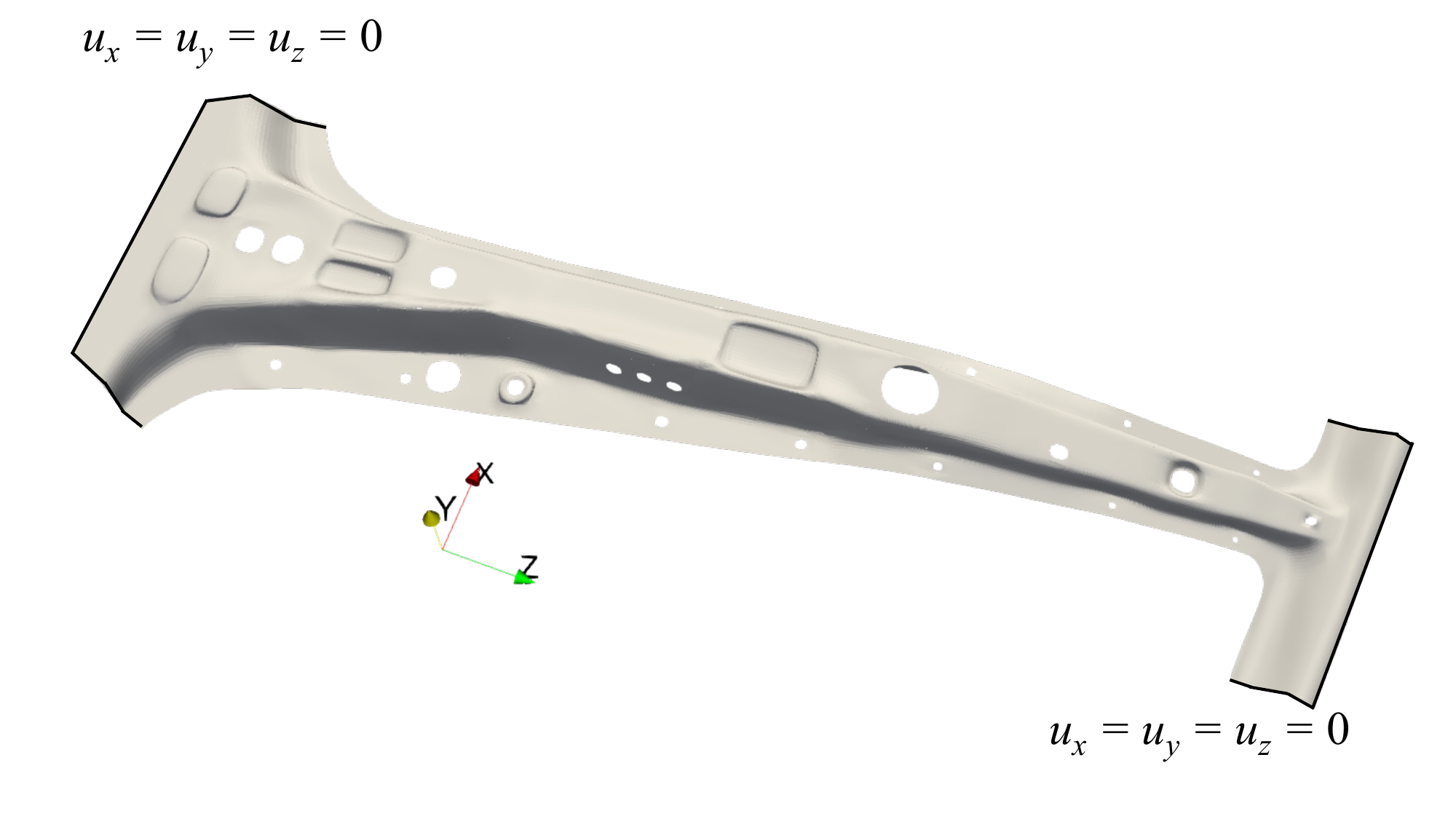} 
\caption{B-pillar model with two ends fully clamped, subjected to $y$-direction loading over the entire surface.} \label{fig:bpillar_model}
\end{figure}

\begin{figure}[tbp] 
\centering 
\includegraphics[width=\linewidth]{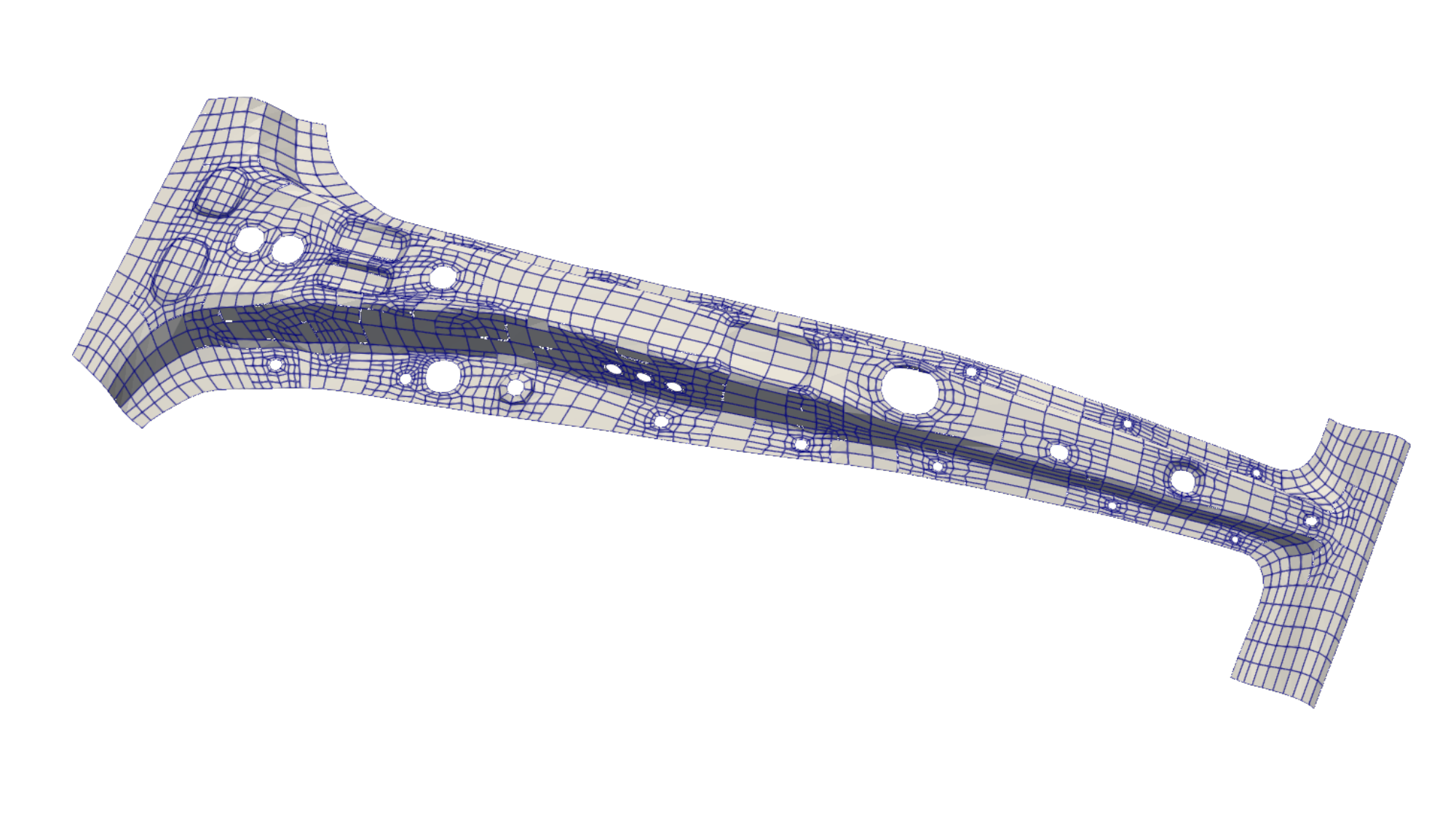} 
\caption{B\'ezier mesh of the B-pillar, consisting of 2881 elements and 3507 control points.} 
\label{fig:bpillar_bzmesh} 
\end{figure}

We next investigate O-SNet on a complex engineering structure, the B-pillar represented by ASUTS. As shown in Fig.~\ref{fig:bpillar_model}, both ends of the structure are fully clamped. Distributed surface loadings are applied in the $y$-direction. The corresponding B\'ezier mesh is shown in Fig.~\ref{fig:bpillar_bzmesh}.

A total of 1250 loading functions are generated by GRF using an amplitude factor of $1.5 \times 10^7$, with 1000 cases for training and 250 cases for testing. The branch net is an MLP with four residual blocks~\cite{he2016deep}. We also compare three network widths with 250, 500, and 750 neurons per hidden layer.

\begin{figure}[tbp]
    \centering

    \begin{subfigure}[t]{0.8\linewidth}
        \centering
        \includegraphics[page=3, width=\linewidth]{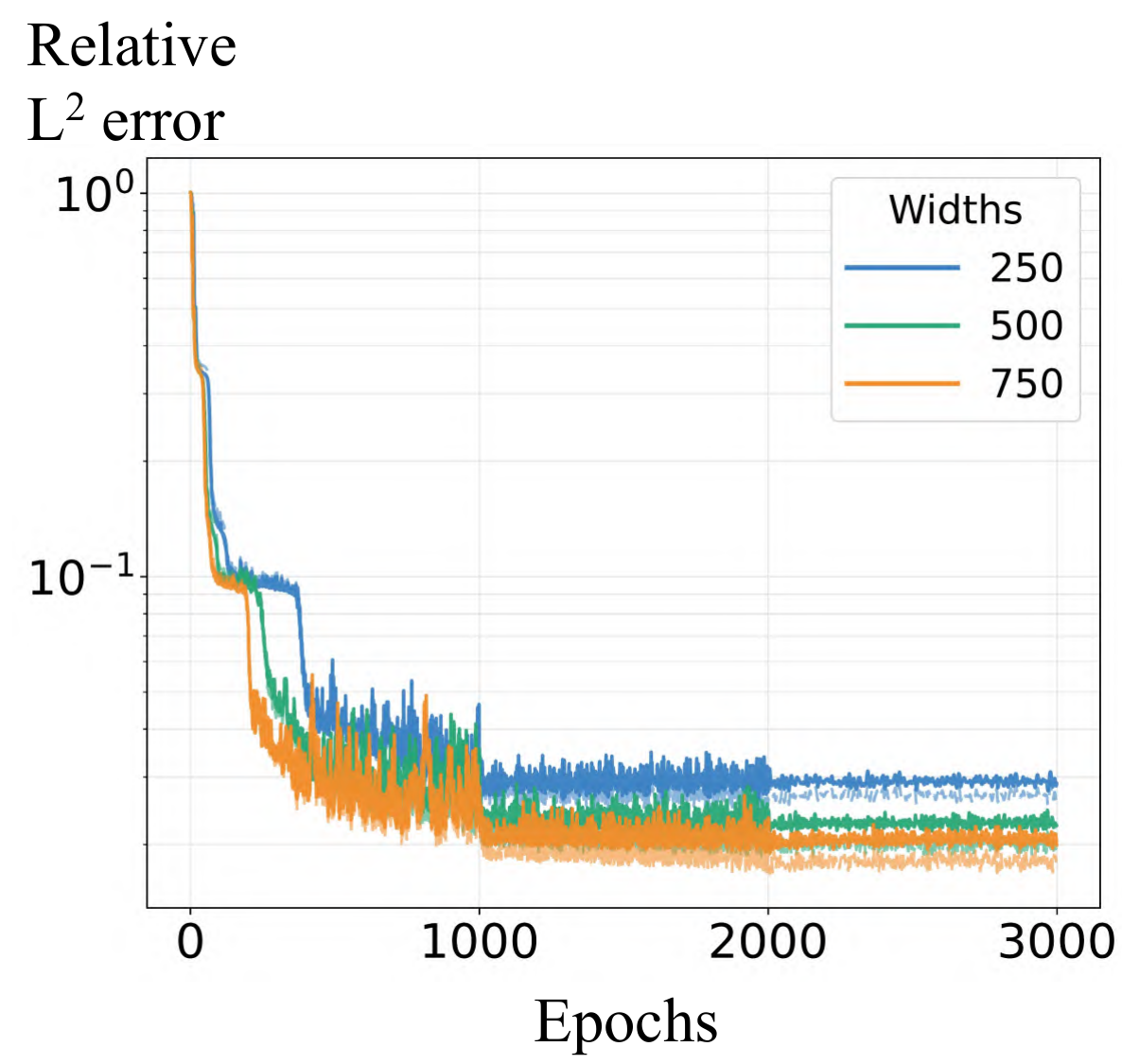}
        \caption{}
        \label{fig:bpillar L2 a}
    \end{subfigure}
    \hfill
    \begin{subfigure}[t]{0.8\linewidth}
        \centering
        \includegraphics[page=4, width=\linewidth]{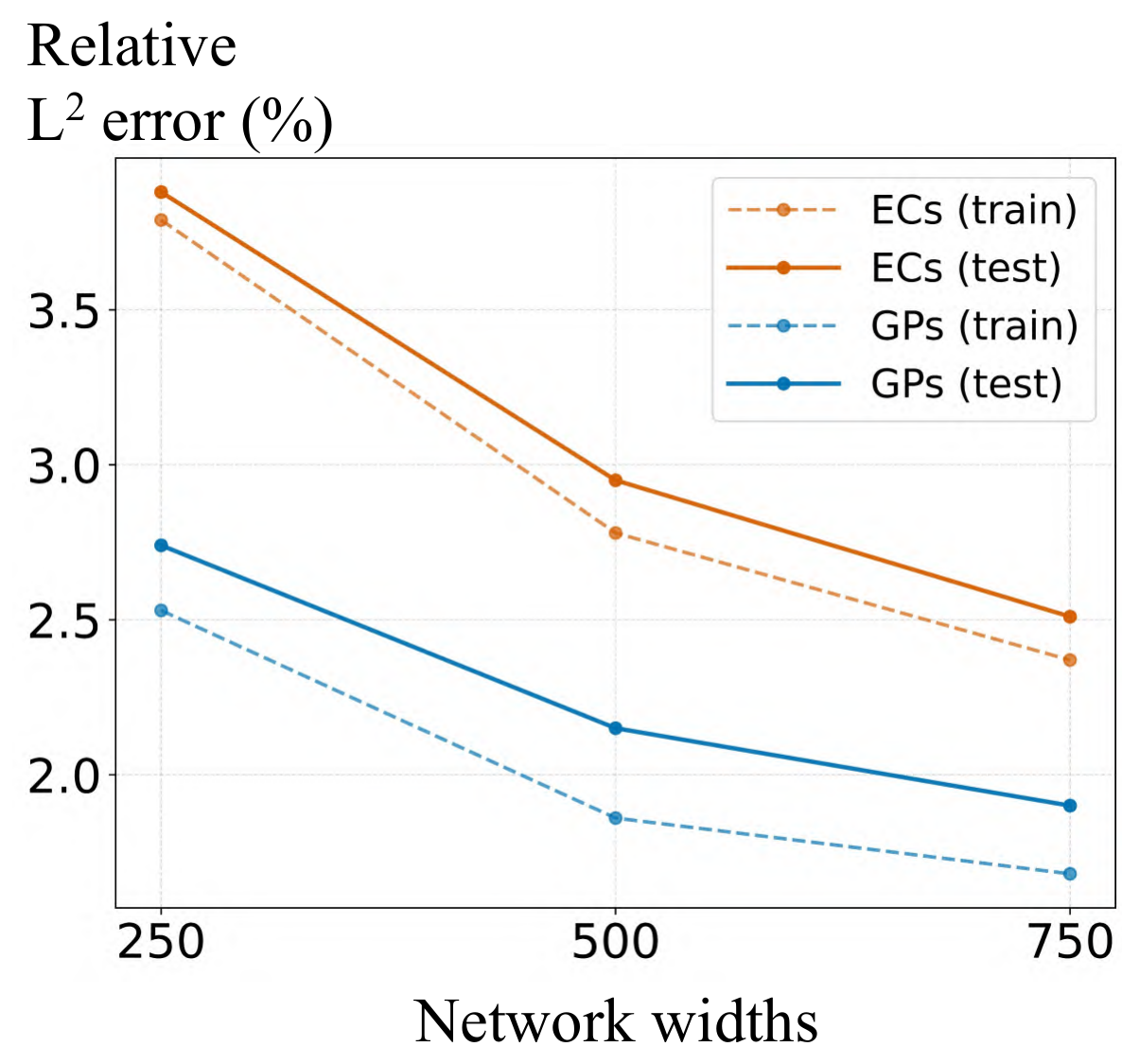}
        \caption{}
        \label{fig:bpillar L2 b}
    \end{subfigure}

    \caption{Training curve and the network width comparison for the B-pillar problem. 
    (a) Mean relative \(L^2\) error versus epochs, where solid and dashed lines denote test and training errors, respectively (with learning rates \(10^{-3}\), \(10^{-4}\), and \(10^{-5}\), each used for 1000 epochs). 
    (b) Minimum relative \(L^2\) errors for different network widths, evaluated at GPs and ECs.}
    \label{fig:bpillar L2}
\end{figure}

Fig.~\ref{fig:bpillar L2} summarizes the training curves and the influence of network width. The mean relative $L^2$ error of the $y$-displacement is evaluated on both training and test sets at GPs and ECs. The 750-neuron network is selected, and its average errors are reported in Table~\ref{tab:bpillar_error}.

\begin{table}[tbp]
\caption{Average relative errors of the $y$-direction displacement in the B-pillar problem with 750 neurons.}
\centering
\begin{tabular*}{\linewidth}{@{\extracolsep{\fill}}ccc@{}}
\hline
Error type & Training set & Test set \\
\hline
GPs & 1.68\% & 1.90\% \\
ECs & 2.38\% & 2.51\% \\
\hline
\end{tabular*}
\label{tab:bpillar_error}
\end{table}

\begin{figure*}[p]
    \centering
    \begin{minipage}{0.8\linewidth}
    \centering
    \includegraphics[width=\linewidth,keepaspectratio]{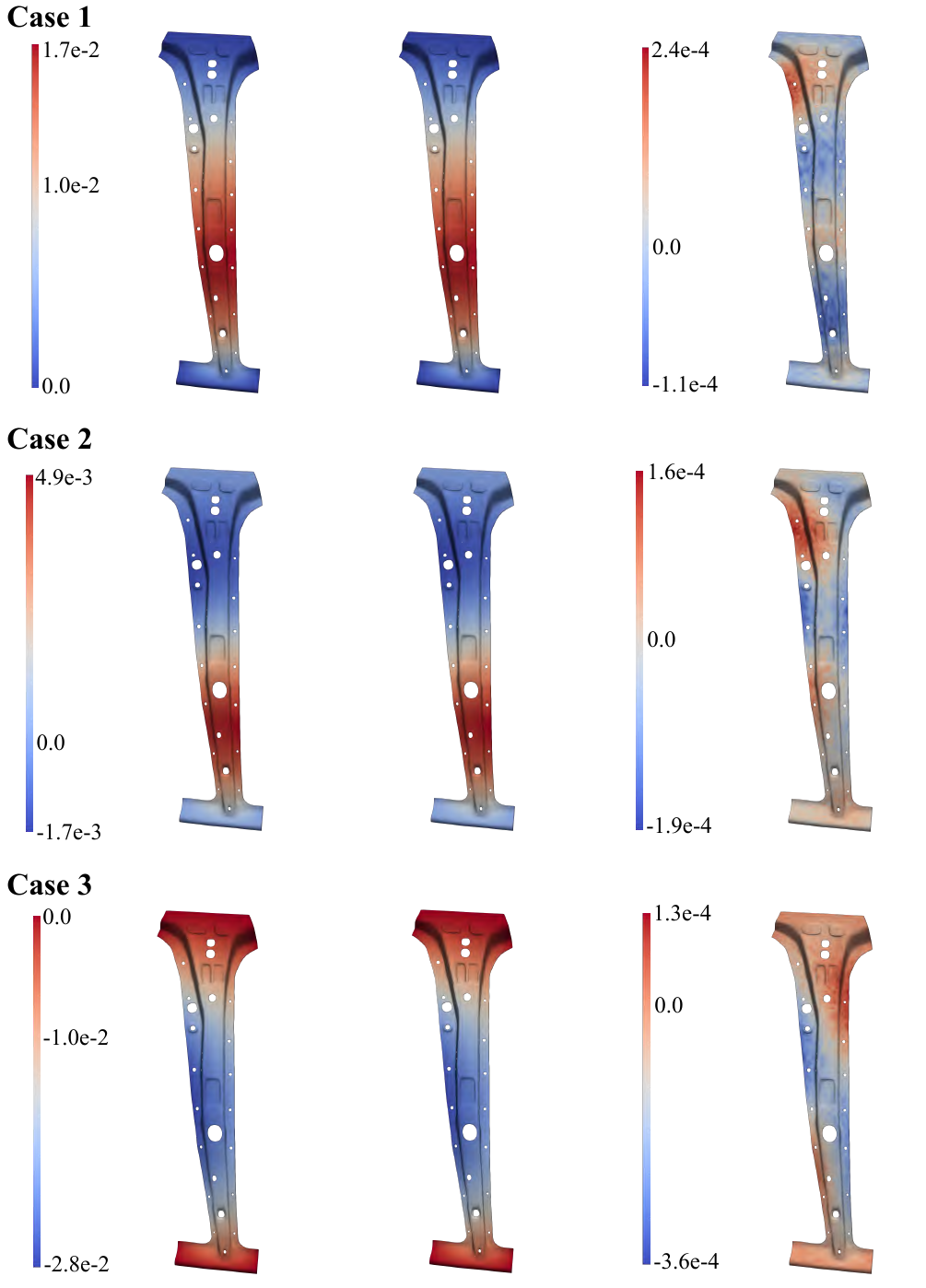}

    \vspace{0.3em}
    
    \makebox[\linewidth][l]{%
        \hspace*{0.22\linewidth}\makebox[0pt][c]{(a) IGA result}%
        \hspace*{0.30\linewidth}\makebox[0pt][c]{(b) Prediction by O-SNet}%
        \hspace*{0.36\linewidth}\makebox[0pt][c]{(c) Error}%
    }

    \end{minipage}

    \vspace{0.3em}
    \caption{Comparison of $y$-displacement fields for three representative test cases of the B-pillar problem.}
    \label{fig:bpillar cases}
\end{figure*}

Representative cases (the first, middle, and last sample in the test dataset) are shown in Fig.~\ref{fig:bpillar cases}. The relative $L^2$ errors at GPs and ECs are $0.70\%$ and $1.08\%$ for Case 1, $2.55\%$ and $4.05\%$ for Case 2, and $0.76\%$ and $0.93\%$ for Case 3, respectively. These results show that O-SNet maintains accurate predictions for a large-scale engineering shell structure with complex geometry.

\FloatBarrier
\subsubsection{Geometrically nonlinear case: plane nose}

In the last case, we study O-SNet on a real-world engineering structure, the plane nose. This example is governed by the geometrically nonlinear Kirchhoff--Love shell formulation~\cite{kiendl2009isogeometric,kiendl2015isogeometric}, which introduces nonlinear load--displacement behavior and thus poses a more challenging operator learning task.

\begin{figure}[tbp]
    \centering
    \includegraphics[width=\linewidth]{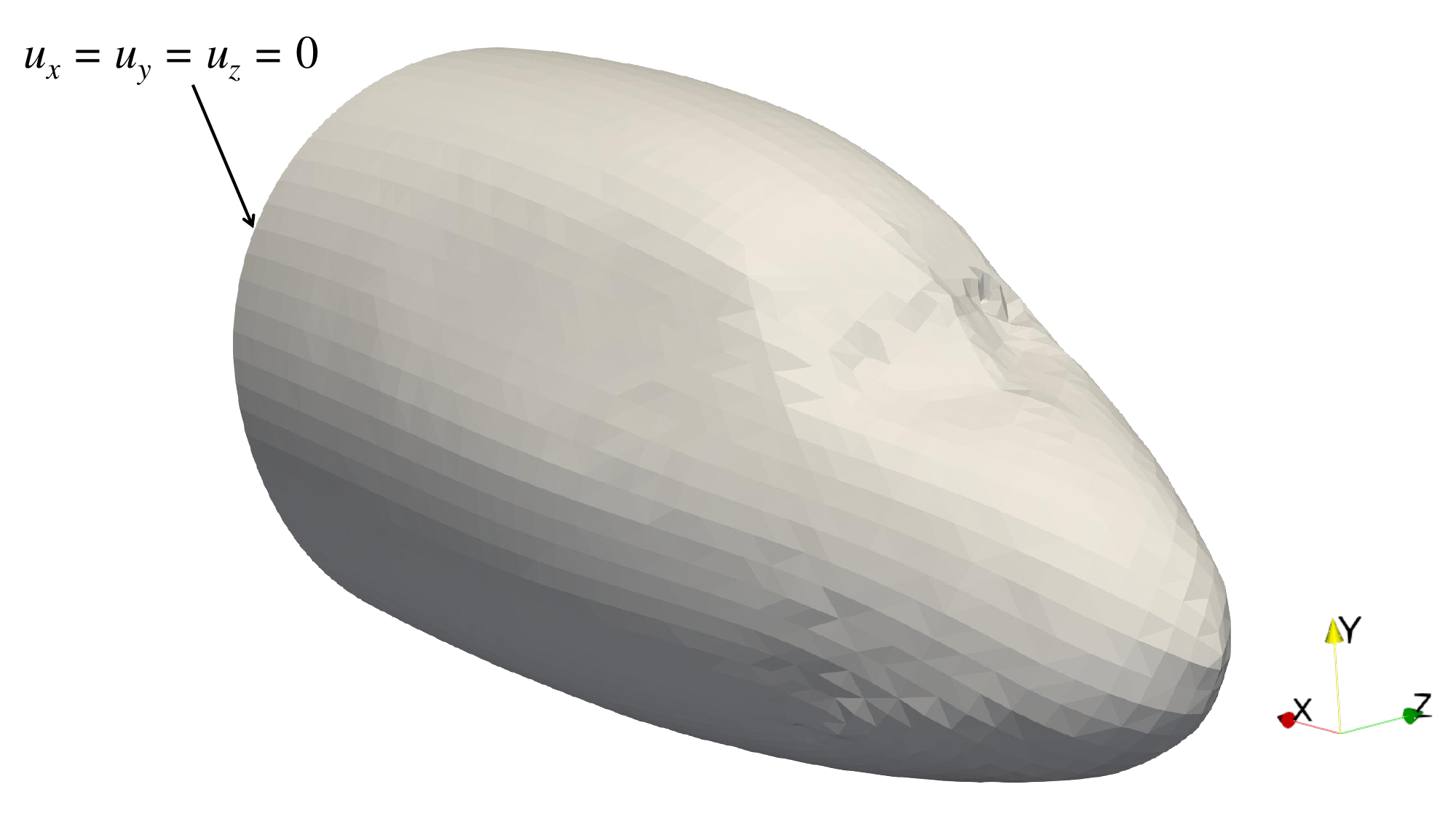}
    \caption{Plane nose model with the left circular edge fully clamped, subjected to $x$-direction loading over the entire surface.}
    \label{fig:head_model}
\end{figure}

\begin{figure}[tbp]
    \centering
    \includegraphics[width=\linewidth]{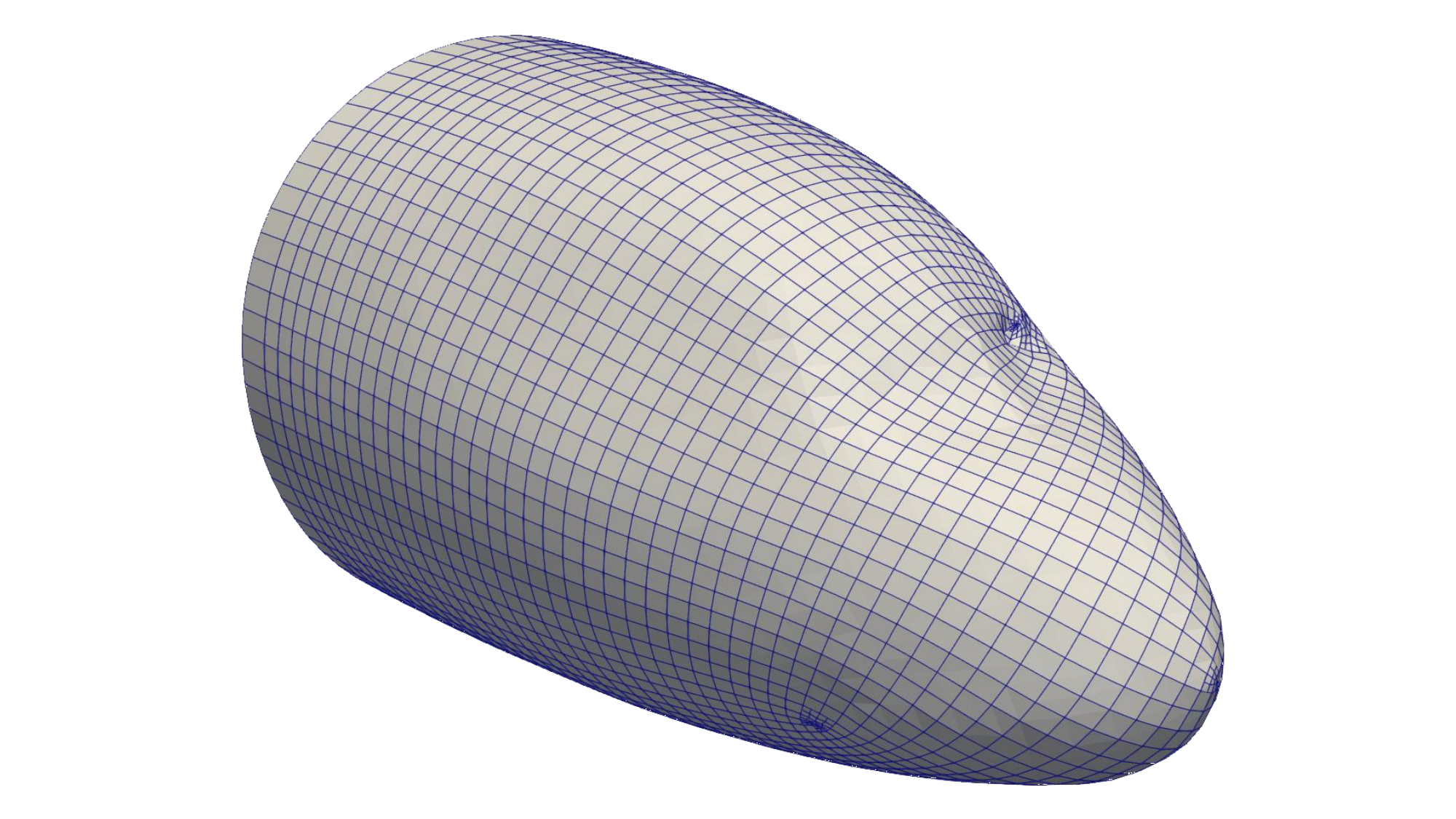}
    \caption{B\'ezier mesh of the plane nose, consisting of 3011 elements and 3141 control points.}
    \label{fig:head_bzmesh}
\end{figure}

As shown in Fig.~\ref{fig:head_model}, the circular edge on the left side, corresponding to the $yz$-plane cut, is fully clamped. Distributed surface loadings are applied over the entire surface in the $x$-direction. The corresponding B\'ezier mesh is shown in Fig.~\ref{fig:head_bzmesh}.

Similar to the roof and B-pillar cases, a total of 1250 loading functions are generated using GRF with an amplitude factor of $5 \times 10^8$. The dataset is split into 1000 training cases and 250 test cases. The displacement responses are obtained from the isogeometric solver tIGAr~\cite{kamensky2019tigar}. Unlike the linear Kirchhoff--Love shell cases, the plane nose uses $E=2.06\times 10^{11}$, $\nu=0.3$, and $t=0.03$.

The branch net adopts an MLP with four residual blocks. We compare three network widths with 250, 500, and 750 neurons per hidden layer. 

\begin{figure}[tbp]
    \centering

    \begin{subfigure}[t]{0.8\linewidth}
        \centering
        \includegraphics[page=5, width=\linewidth]{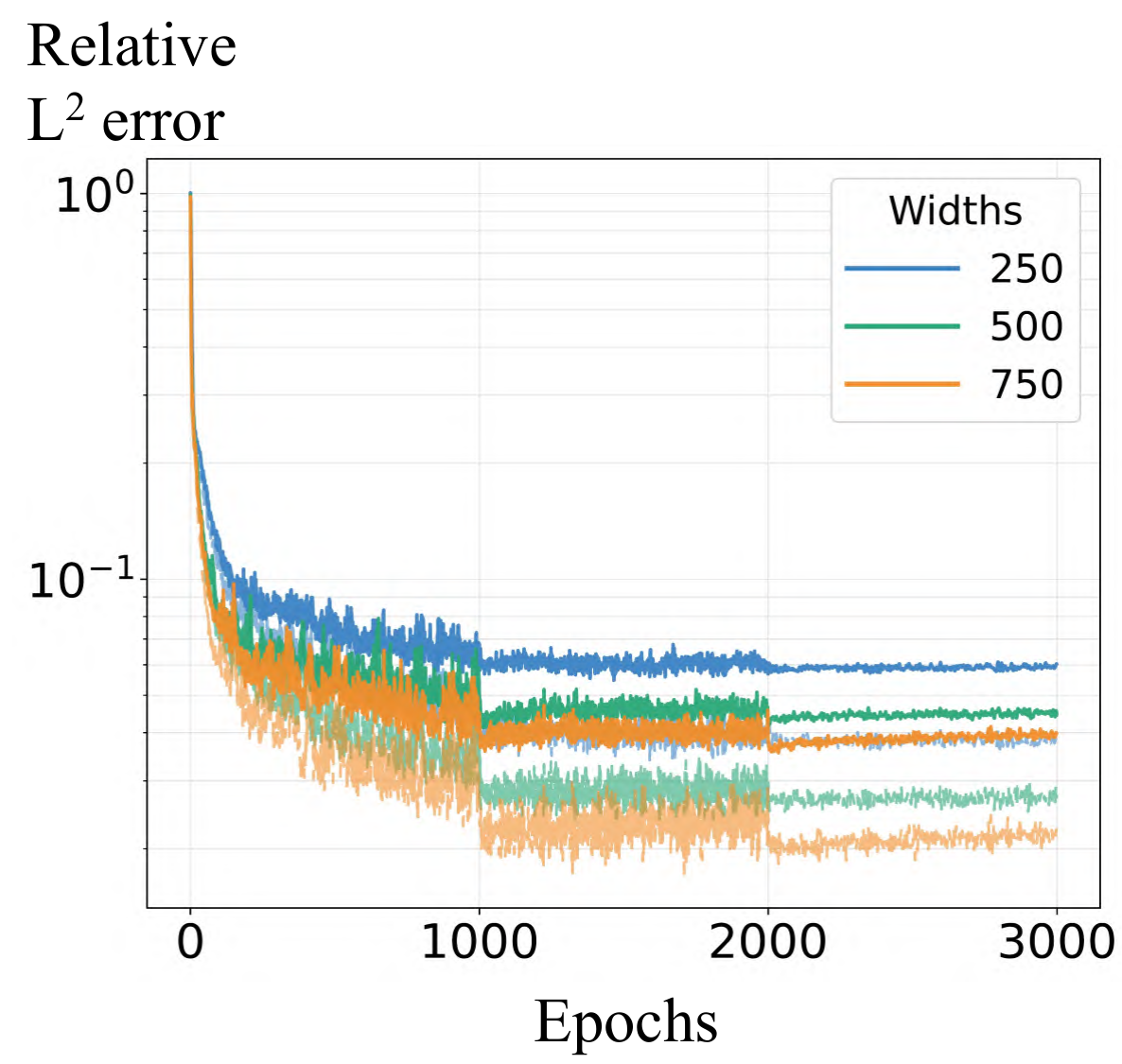}
        \caption{}
        \label{fig:plane L2 a}
    \end{subfigure}
    \hfill
    \begin{subfigure}[t]{0.8\linewidth}
        \centering
        \includegraphics[page=6, width=\linewidth]{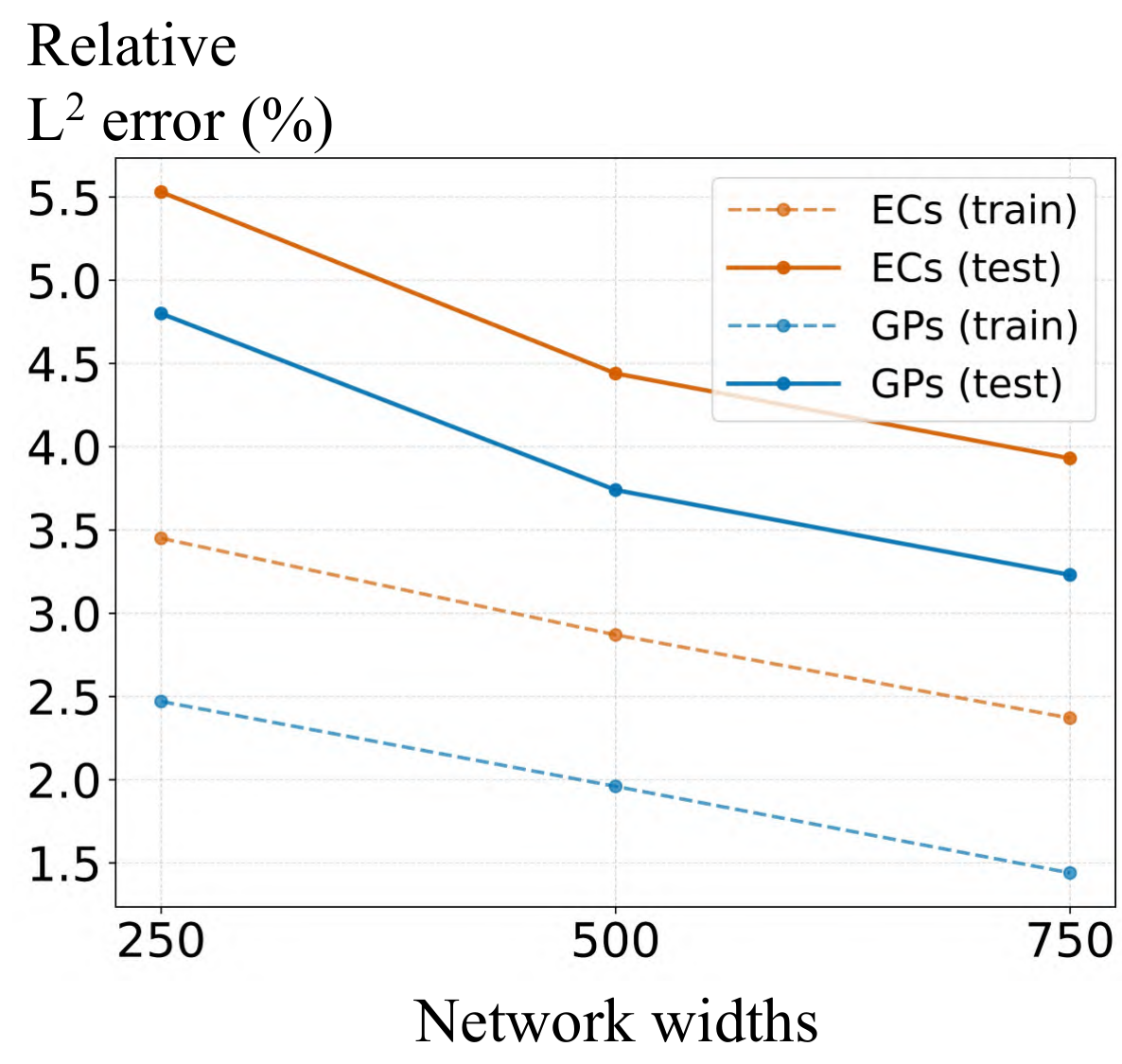}
        \caption{}
        \label{fig:plane L2 b}
    \end{subfigure}

    \caption{Training curve and the network width comparison for the geometrically nonlinear plane nose problem. 
    (a) Mean relative \(L^2\) error versus epochs, where solid and dashed lines denote test and training errors, respectively (with learning rates \(10^{-3}\), \(10^{-4}\), and \(10^{-5}\), each used for 1000 epochs).
    (b) Minimum relative \(L^2\) errors for different network widths, evaluated at GPs and ECs.}
    \label{fig:head nonlinear L2}
\end{figure}

\clearpage
\begin{strip}
    \centering

    \begin{minipage}[t]{0.48\textwidth}
    \vspace{0pt}
    Fig.~\ref{fig:head nonlinear L2} summarizes the training curves and the influence of the network width for the nonlinear case. The errors are evaluated at both GPs and ECs. The average errors of the best case are reported in Table~\ref{tab:head_error}.
    \end{minipage}\hfill
    \begin{minipage}[t]{0.48\textwidth}
    \vspace{0pt}
    \refstepcounter{table}\label{tab:head_error}
    \noindent\textbf{Table~\thetable:} Average relative errors of the $x$-direction displacement in the plane nose problem with 750 neurons.

    \vspace{0.2em}
    \centering
    \begin{tabular*}{\linewidth}{@{\extracolsep{\fill}}ccc@{}}
    \hline
    Error type & Training set & Test set \\
    \hline
    GPs & 1.44\% & 3.23\% \\
    ECs & 2.37\% & 3.93\% \\
    \hline
    \end{tabular*}
    \end{minipage}

    \vspace{1.0em}

    \begin{minipage}{\textwidth}
    \centering
    \includegraphics[width=\linewidth,keepaspectratio]{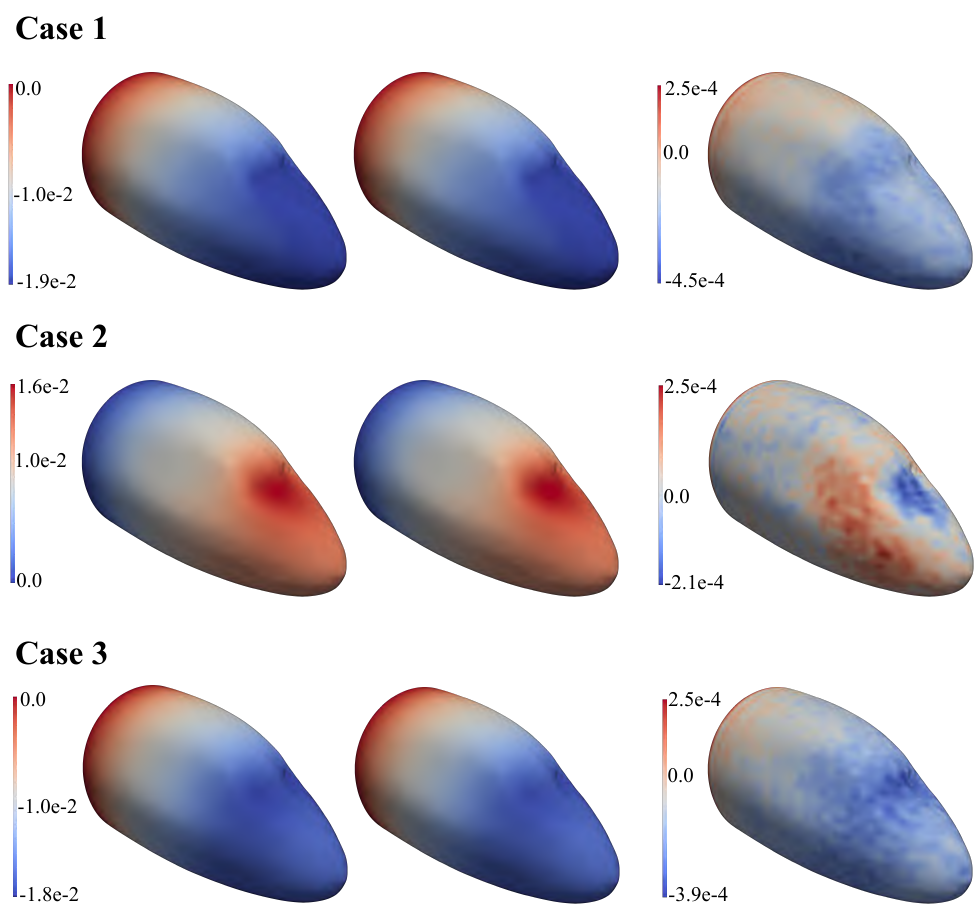}

    \vspace{0.3em}
    
    \makebox[\linewidth][l]{%
        \hspace*{0.22\linewidth}\makebox[0pt][c]{(a) IGA result}%
        \hspace*{0.29\linewidth}\makebox[0pt][c]{(b) Prediction by O-SNet}%
        \hspace*{0.35\linewidth}\makebox[0pt][c]{(c) Error}%
    }

    \end{minipage}

    \vspace{0.3em}
    \captionof{figure}{Comparison of $x$-displacement fields for three representative test cases of the plane nose problem.}
    \label{fig:head cases}
\end{strip}

Representative cases (the first, middle, and last sample in the test dataset) are presented in Fig.~\ref{fig:head cases}. The relative $L^2$ errors at GPs and ECs are $1.94\%$ and $2.00\%$ for Case 1, $0.70\%$ and $1.11\%$ for Case 2, and $1.55\%$ and $1.64\%$ for Case 3, respectively. These results demonstrate that O-SNet can learn the load-response operator for geometrically nonlinear shell problems with complex geometries.

\section{Conclusion and future work}
\label{section 6}
In this work, we propose a novel isogeometric deep learning method, SplineNet, which unifies scientific deep learning with CAD/CAE integration. B\'ezier extraction provides a unified mechanism for embedding ASUTS into neural networks. SplineNet has been applied to both solution learning (P-SNet) and operator learning (O-SNet).

We first verified P-SNet on the Scordelis--Lo roof benchmark by employing the KL shell energy to solve a single-instance problem. This physics-informed version highlights the feasibility of using a spline-based neural representation for shell analysis. Building on this foundation, we further introduced O-SNet, where the displacement fields are instantly predicted given unseen loading functions for complex geometries. In contrast to sampling-based operator learning methods, our method operates on control variables.

In the future, a particularly promising direction is to explore the proposed method in gradient-based shape optimization, where the position of control points can be treated as design variables within a CAD/CAE-integrated loop. Extending the approach to more general nonlinear settings (e.g., advanced constitutive models, contact, and buckling) and improving computational efficiency will further enhance its applicability to practical engineering design.

\bmhead{Acknowledgments}
S.~Luo and X.~Wei are partially supported by National Natural Science Foundation of China (No.~12494550/12494555 and No.~12571408).

\FloatBarrier
\bibliography{reference}

\end{document}